\documentclass[twoside,11pt]{article}

\usepackage{jmlr2e}
\usepackage{amsfonts}
\usepackage{amsmath}
\usepackage{multicol}
\usepackage{booktabs}

\usepackage{multirow}
\usepackage{sidecap}
\usepackage[dvipsnames]{xcolor}
\usepackage{hyperref}
\hypersetup{
    colorlinks=True,
    citecolor=Green,
    pdfborder={0 0 0},
}

\firstpageno{1}
\iftrue
\fi

\newif\ifLongVersion
\LongVersionfalse

\usepackage{babel}

\usepackage{subcaption}
 \newtheorem{assumption}{Assumption}

\usepackage{cleveref}

\usepackage{lastpage}

\jmlrheading{11}{2024}{1-\pageref{LastPage}}{11/24; Revised --/--}{--/--}{21-0000}{Wang, Shen, Tao, Sun, Zheng and Tao}

\ShortHeadings{Sample JMLR Paper}{A Unified Analysis for Finite Weight Averaging}
\firstpageno{1}

\begin{document}

\title{A Unified Analysis for Finite Weight Averaging}

\author{\name Peng Wang $^1$$^{,}$$^5$ \email wang\_peng\_2020@163.com \\
       \name Li Shen $^2$\thanks{Corresponding author} \email mathshenli@gmail.com \\
       \name Zerui Tao $^3$ \email zerui.tao@riken.jp \\
       \name Yan Sun $^4$ \email ysun9899@uni.sydney.edu.au \\
       \name Guodong Zheng $^1$ \email gd\_zheng@hust.edu.cn \\
       \name Dacheng Tao $^5$ \email dacheng.tao@ntu.edu.sg \\       
       \addr $^1$ School of Mathematics and Statistics, Huazhong University of Science and Technology, Wuhan, China \\
       \addr $^2$ School of Cyber Science \& Technology, Shenzhen Campus of Sun Yat-sen University, Shenzhen, China \\
       \addr $^3$ RIKEN Center for Advanced Intelligence Project,  Tokyo, Japan \\
       \addr $^4$ Faculty of Engineering, the University of Sydney, Australia \\
       \addr $^5$ School of Computer and Data Science, Nanyang Technological University, Singapore}
       %

\editor{My editor}

\maketitle

\begin{abstract}
Averaging iterations of Stochastic Gradient Descent (SGD) have achieved empirical success in training deep learning models, such as Stochastic Weight Averaging (SWA), Exponential Moving Average (EMA), and LAtest Weight Averaging (LAWA). Especially, with a finite weight averaging method, LAWA can attain faster convergence and better generalization. However, its theoretical explanation is still less explored since there are fundamental differences between finite and infinite settings. 
In this work, we first generalize SGD and LAWA as Finite Weight Averaging (FWA) and explain their advantages compared to SGD from the perspective of optimization and generalization.
A key challenge is the inapplicability of traditional methods in the sense of expectation or optimal values for infinite-dimensional settings in analyzing FWA's convergence. Second, the cumulative gradients introduced by FWA introduce additional confusion to the generalization analysis, especially making it more difficult to discuss them under different assumptions.
Extending the final iteration convergence analysis to the FWA, this paper, under a convexity assumption, establishes a convergence bound $\mathcal{O}(\log\left(\frac{T}{k}\right)/\sqrt{T})$, where $k\in[1, T/2]$ is a constant representing the last $k$ iterations. Compared to SGD with $\mathcal{O}(\log(T)/\sqrt{T})$, we prove theoretically that FWA has a faster convergence rate and explain the effect of the number of average points. 
In the generalization analysis, we find a recursive representation for bounding the cumulative gradient using mathematical induction. We provide bounds for constant and decay learning rates and the convex and non-convex cases to show the good generalization performance of FWA.
Finally, experimental results on several benchmarks verify our theoretical results. 
\end{abstract}

\begin{keywords}
Finite weight averaging, Generalization bound, Last iteration convergence.
\end{keywords}

\section{Introduction}
Weight averaging methods have shown great success in diverse applications ranging from convex to non-convex optimization. From the perspective of geometric intuition, a converging optimizer fluctuates around a minimum value, so weight-averaging methods tend to approach the minimum.
In the 90s, \citet{polyak1992acceleration} have proven that averaging the model weights can accelerate the convergence speed in the convex loss function regime. Now, in deep learning,  the prevalence of large models often results in high-dimensional loss functions and non-convex, which lead to the creation of many variants of averaging methods, such as Exponential Moving Average (EMA) \citep{szegedy2016rethinking}, Stochastic Weight Averaging (SWA) \citep{izmailov2018averaging}, LAtest Weight Averaging (LAWA) \citep{kaddour2022stop,sanyal2023early}. These methods are widely used for training foundation models and have been empirically shown to perform well. The most commonly held geometric intuition is that all the averaging techniques aim to achieve a flat landscape, which yields a better generalization.

SWA, averaging weights with the same loss from pre-trained models to move them to better generalizing regions. Unlike the former, EMA assigns higher weights to the more recent checkpoints. SWA and EMA, the most commonly used enhancing generalization in practice, rely on historical averaging results in their update rules. However, \citet{kaddour2022stop} shows that premature use of historical information can result in suboptimal results, hurting its performance and the experiment with an exponential moving average worse than uniform coefficients. Trainable Weight Averaging (TWA) \citep{li2022trainable} achieves accelerated convergence and improved generalization by learning the averaging coefficients, but it requires gradient projection in each step. As shown in Eq.~\eqref{LAWA-rules}, LAWA averages the $k$ checkpoints collected by the end of each epoch and has been empirically observed to improve both convergence and generalization in pre-training large language models \citep{sanyal2023early}. It avoids additional computational costs and has good performance in terms of convergence and generalization, which sparks our intense research interest.

 FWA collects finite points and averages at the end of the training, which is also called "tail-averaging", and has been studied before in \cite{jain2017markov,jain2018accelerating,jain2018parallelizing} where prove the benefits of tail-averaging for linear regression. This paper provides a theoretical exploration of FWA, focusing on optimization and generalization. Based on the mathematical definition presented in Eq.~\eqref{FWA-rules}, FWA connects SGD and LAWA precisely. LAWA is obtained by collecting points with intervals in finite iterations, and FWA degenerates into SGD when $k=1$. Therefore, a comparative analysis of the theoretical advantages of a class of finite-point averaging methods in terms of optimization and generalization is presented.

\begin{table*}[t]
\caption{\small Comparison of FWA with SGD on different settings.
Here $T$ represents iterations, $n$ denotes the size of datasets, $k, c$, and $\beta$ are constants, $\rho_i$ is a weighted parameter. We can derive FWA has sharper bounds compared to SGD in the different settings, particularly in the Corollary \ref{thm:stability-con-with-cor} and \ref{thm:stability-non-with} when $k=1$, where all bounds are degraded to SGD.}
\label{sample-table1}
\vspace{-0.2cm}
\begin{center}
\renewcommand\arraystretch{1.3}
\resizebox{\textwidth}{!}{
\begin{sc}
\small
\begin{tabular}{l|c|c|c}
\toprule
Settings & Algorithm & learning rate & Generalization Bound  \\ \midrule 
\multirow{4}{*}{convex} & SGD & $\alpha_i = \alpha (Constant)$ & $\frac{2\alpha L^2 T}{n}$ \cite{hardt2016train}\\
& SWA & $\alpha_i = \alpha (Constant)$ & $\frac{\alpha L^2 T}{n}$ \cite{hardt2016train,xiao2022stability,pmlr-v235-wang24bl}\\
&\color{blue} FWA & \color{blue} $\alpha_i = \alpha, \rho_i = 1$ & \textcolor{blue}{$\frac{2\alpha L^2}{n} \left( T - \frac{k}{2} \right)$} COR: \ref{thm:stability-con-with-cor} \\ 
&\color{blue} FWA & \color{blue} $\alpha_i, \rho_i$ & \textcolor{blue}{$\frac{2L^2}{nk} \left( \sum_{t=1}^{k}\sum_{i=1}^{t} \rho_i \alpha_i + \sum_{t=k+1}^{T}\sum_{i=t-k+1}^{t} \rho_{i-(t-k)} \alpha_i \right)$} THM: \ref{thm:stability-con-with} \\ \hline
\multirow{4}{*}{non-convex} &  SGD & $\alpha_t = \frac{c}{t}$ (Decay) & $\mathcal{O}(T^{\frac{c\beta}{1+c\beta}}/n)$ \cite{hardt2016train}\\ 
&  SWA & $\alpha_t = \frac{c}{T}$ (Constant) & $\mathcal{O}(T^{\frac{c\beta}{2+c\beta}}/n)$ \cite{pmlr-v235-wang24bl}\\ 
& \color{blue} FWA & \color{blue} $\alpha_t=\alpha=\frac{c}{T}, \rho_i=1$ & \textcolor{blue}{$\mathcal{O}\left(T^{\frac{c\beta}{c\beta+k}}/n\right)$} THM: \ref{thm:stability-non-with}\\
& \color{blue} FWA & \color{blue} $\alpha_t=\frac{c}{t}, \rho_i$ & \textcolor{blue}{$\mathcal{O}\left(T^{\frac{kc\beta+c^2 \beta^2}{2kc\beta+c^2 \beta^2 +k^2 (1-c\beta)}}/n \right)$} THM: \ref{thm:stability-non-with-decay}\\ 
\bottomrule
\end{tabular}
\end{sc}
}
\end{center}
\vskip -0.2in
\end{table*}

\begin{table*}[t]
\caption{\small Comparison of FWA with SGD and LAWA on optimization.
Here $T$ represents iterations, $k$ denotes the last $k$ iterations, and $d$ is the interval length. We can derive that FWA has a faster convergence rate than SGD and LAWA.}
\label{sample-table2}
\vspace{-0.2cm}
\begin{center}
\renewcommand\arraystretch{1.3}
\resizebox{\textwidth}{!}{
\begin{sc}
\small
\begin{tabular}{lccc}
\toprule
 Algorithm & SGD & FWA & LAWA\\ \midrule
Output & Last & Finite AVE & Finite AVE\\ \hline
Assumptions & Convex \& Non-smooth & Convex \& Non-smooth & Convex \& Non-smooth \\ \hline
Learning Rate & $\frac{c}{\sqrt{t}}$ & $\frac{c}{\sqrt{t}}$ & $\frac{c}{\sqrt{t}}$ \\ \hline
Convergence Rate & $\mathcal{O}(\log(T)/\sqrt{T})$ \cite{shamir2013stochastic} & \textcolor{blue}{$\mathcal{O}(\log(\frac{T}{2k})/\sqrt{T})$} THM: \ref{thm:convergence1} & \textcolor{blue}{$\mathcal{O}(d\log(\frac{T}{2kd})/\sqrt{T})$} THM: \ref{thm:convergence-LAWA}\\ \hline

\bottomrule
\end{tabular}
\end{sc}
}
\end{center}
\vskip -0.2in
\end{table*}

Exploring the theoretical basis for FWA's superiority over SGD primarily faces the following challenges. In optimization, classical convergence analysis techniques typically depend on information from all iterations, resulting in slower convergence rates that often do not align well with practical training speeds. It certainly does not apply to the study of FWA either. The convergence of the final iteration, which exhibits improved convergence, has garnered significant research interest. However, as these techniques are still emerging and evolving, their application across various settings remains an open challenge \citep{kaddour2022stop}. 
In generalization, building the boundary of SGD for the step $T$ only requires binding the gradient generated by the $T$-th sample.
However, the $k$ historical information before the $T$-th step should be bounded for FWA. For non-convex functions, where $(1+\alpha\beta)$-expensive is used for updating and different learning rates are considered, which makes the theoretical analysis even more complex and leads to varying results. In particular, decaying learning rates complicate the computation of accumulated gradients and the bound, making the process notably more challenging.

To mitigate these theoretical deficiencies, we construct the generalization bound using mathematical induction, enabling us to derive a recursive representation that bounds the gradient at each step. Then, the problem of bounding each step is transformed into a problem of bounding a finite sum, which solves the difficulties in non-convex settings. For the complex expressions arising from the decaying learning rate, we derive explicit expressions for the upper bounds of the cumulative gradient through mathematical analysis. Our results, under the convex case, suggest that FWA can improve the generalization bound in various settings compared with SGD. For the non-convex case, the generalization bound of FWA with a decaying learning rate is larger than that of SGD due to its larger accumulated gradients. However, when a smaller constant learning rate is used, the generalization bound of FWA can be significantly smaller than that of SGD, demonstrating that FWA has the potential to enhance generalization performance (see Table \ref{sample-table1}). In addition, our main result in convergence is an $\mathcal{O}(\log(T/2k)/\sqrt{T}), 1\leq k\leq T/2,$ upper bound for FWA minimizing convex functions with a decay step size schedule $\alpha_t = \frac{c}{\sqrt{t}}$ whether smooth or not, generalizing the result in \citep{shamir2013stochastic}, where the focus is on the strongly convex functions. It surpasses the bound $\mathcal{O}(\log(T)/\sqrt{T})$ of SGD under identical assumptions, which also implies that, under the constraint of $1\leq k\leq T/2$, the larger $k$ is, the faster the convergence (see Table \ref{sample-table2}). 

\subsection{Contributions}

This paper mainly focuses on theoretical exploration, and some experimental results are intended to verify its correctness.
Specifically, we illustrate that FWA can achieve better generalization and faster convergence than SGD. Table \ref{sample-table1} and Table \ref{sample-table2} summarize the main theoretical results. Our contributions are listed as follows.

\begin{itemize}
\item Our first theoretical investigation focuses on algorithms such as LAWA, and we provide a theoretical explanation for its more general form, FWA. We elucidate why it exhibits better generalization and faster convergence compared to SGD.

\item We derive stability-based generalization bounds for FWA in the various settings -- convex or non-convex, constant or decay learning rate -- for showing the good generalization performance of FWA and prove that the larger $k$ is, the smaller the bounds (see Table \ref{sample-table1}). Mainly, when the objective function is non-convex, we find that the generalization bound of FWA with a decaying learning rate is not always smaller than that of SGD. However, under smaller learning rates, FWA significantly improves generalization, achieving a generalization bound that is markedly tighter than that of SGD. 

\item We obtain a convergence rate for FWA under the convex and non-smooth functions. Based on this, we provide a theoretical perspective on the convergence of FWA is faster than the SGD and prove that under the constraint of $1\leq k\leq T/2$, the larger $k$ is, the faster the convergence. Moreover, FWA performs better than LAWA in the same setting where the larger the interval length is, the slower the convergence (see Table \ref{sample-table2}). 

\item We provide experiments for FWA with or without replacement sampling to verify our results based on the metric parameter distance and generalization error on the CIFAR10 and Adult datasets, respectively. The experiments also coincide with our theoretical findings.
\end{itemize}

\section{Related Work}\label{related work}

\textbf{Averaging algorithm.} Weight averaging has wildly shown advantages in generalization and convergence speed under simple linear \citep{lakshminarayanan2018linear} and convex settings \citep{ruppert1988efficient,polyak1992acceleration}. Recently, this idea has been extended to train deep learning models \citep{garipov2018loss,sanyal2023early}. SWA \citep{izmailov2018averaging} averages the checkpoints along the training trajectories of SGD. A similar approach is the Stochastic Weight Averaging Densely \citep{cha2021swad}, which averages after collecting weight densely and empirically shows flatter minima can be found. Another approach that focuses on weighted averages, such as EMA \citep{szegedy2016rethinking}, assigns higher weights to the more recent checkpoints. Furthermore, TWA \citep{li2022trainable} employs trainable averaging coefficients to improve the efficiency of SWA further. This paper focuses on a class of finite average algorithms such as LAWA \citep{kaddour2022stop,sanyal2023early}, which average the last collected $k$ latest checkpoints from running a moving window at a predetermined interval. Despite its widespread use and empirical evidence that it achieves faster convergence and better generalization, theoretical analysis is still lacking.

\textbf{Last iteration.} There are many results for the last-iterate convergence of SGD with different assumptions such as convex, non-convex, smoothness, etc. We only focus on convex functions to match our research. An expected convergence $\mathcal{O}(1/T)$ is first obtained for strongly convex functions by \citep{rakhlin2011making}. \citet{shamir2013stochastic} show the first expected last-iterate rates $\mathcal{O}(\log T/\sqrt{T})$ for convex and improves $\mathcal{O}(1/T)$ to $\mathcal{O}(\log T/T)$ for strongly convex objectives. Recent, \citet{jain2021making} improve the previous two rates to $\mathcal{O}(1/\sqrt{T})$ and $\mathcal{O}(1/T)$ under the new step size. Although the bounds of SGD with averaging are universal, the results of finite averaging are rare. A more common form is $\alpha$-suffix $(\alpha \in (0,1))$ \citep{rakhlin2011making,shamir2013stochastic}, but it still depends on $T$ for the average number of last iterations. It has been shown \citep{kaddour2022stop} that having too many averages can harm performance, so we are interested in finite $k$.

\textbf{Stability.} Stability analysis tries to study the generalization ability of an algorithm by its stability \citep{devroye1979distribution,bousquet2002stability,mukherjee2006learning}. Before that, many theoretical tools have been used to explore generalization, such as the VC-dimension \citep{blumer1989learnability,vapnik2006estimation}, the Rademacher complexity \citep{bartlett2002rademacher,koltchinskii2000rademacher}, and the PAC-Bayesian theory \citep{mcallester1999pac}. Specifically, \citet{bousquet2002stability} introduce the concept of algorithm stability from the perspective of statistical learning theory. \citet{hardt2016train} pioneer the use of algorithm stability to study the generalization bounds of SGD, which led to subsequent works by \cite{charles2018stability, zhou2018generalization, yuan2019stagewise, lei2020sharper}. This concept has been applied in various domains, including online learning \citep{yang2021simple}, adversarial training \citep{xiao2022stability}, decentralized learning \citep{zhu2023stability}, and federated learning \citep{sun2023understanding, sun2023mode}. What's more, based on different sampling methods, the stability of the algorithm in the case of sampling without replacement is investigated \citep{shamir2016without,kuzborskij2018data}. Notably, \citet{hardt2016train} and \citet{xiao2022stability} provide generalization and stability analyses of SWA, while their work is constrained to convex functions and sampling with replacement. In this work, we focus on the generalization of FWA under the convex and non-convex settings of both with different learning rates.

\section{Problem Setup}

Let $F(w, z)$ be a loss function that measures the loss of the predicted value of the parameter $w$ at a given sample $z$. There is an unknown distribution $\mathcal{D}$ over examples from some space $\mathcal{Z}$, and a sample dataset $S=(z_1, z_2,..., z_n)$ of $n$ examples i.i.d. drawn from $\mathcal{D}$. Then the \emph{population risk} and \emph{empirical risk} are defined as 
\begin{equation}
R_\mathcal{D}[w]=E_ {z \sim \mathcal{D}}F(w; z) \quad  \textbf{and} \quad 
R_S [w]= \frac{1}{n} \sum_{i=1}^{n}F(w; z_i) .
\end{equation}
 The generalization error of a model $w$ is the difference $\epsilon_{gen}=R_\mathcal{D}[w] - R_S [w]$. Moreover, we mainly introduce the assumptions adopted in the following analysis. 

\begin{assumption}[$L$-Lipschitz]
\label{$L$-Lipschitz}
A differentiable function $F: R^d \rightarrow R$ satisfies the $L$-Lipschitz property, i.e., for $\forall u, v \in R^d, \Vert F(u)-F(v)\Vert \leq L\Vert u-v\Vert$, which implies that $\Vert\nabla F(u)\Vert \leq L$.
\end{assumption}

\begin{assumption}[$\beta$-smooth]
\label{beta-smooth}
A differentiable function $F: R^d \rightarrow R$ is $\beta$-smooth, i.e., for $\forall u, v \in R^d, \Vert\nabla F(u)-\nabla F(v)\Vert \leq \beta\Vert u-v\Vert$, where $\nabla F(v)$ denotes the gradient of $F$ at $v$. 
\end{assumption}

Assumptions \ref{$L$-Lipschitz} and \ref{beta-smooth} are crucial conditions for analyzing the model's generalization. In the analytical process, they are used to establish generalization bounds based on stability.

\begin{assumption}[Convex function]
\label{Convex function}
A differentiable function $F: R^d \rightarrow R$ is convex, i.e., for $\forall u, v \in R^d, F(u) \leq F(v) + \langle\nabla F(u), u-v \rangle.$
\end{assumption}

\begin{assumption}[Bounded variance]
\label{Bounded variance}
The unbiased stochastic gradients $\hat{g}_t=\nabla F_t(w,z)$ of any data sample $z$ satisfies the bounded variance, i.e., for some fixed $G$, $\mathbb{E}[\Vert \hat{g}_t \Vert^2] \leq G^2$.
\end{assumption}

\begin{assumption}[Bounded domain]
\label{Bounded domain}
Assume that the diameter of $W$, namely $\sup_{w,w^{\prime} \in W}\Vert w-w^{\prime} \Vert$ is bounded, i.e., for some constant $D$, $\sup_{w,w^{\prime} \in W}\Vert w-w^{\prime} \Vert \leq D$.
\end{assumption}

Assumption \ref{Bounded variance} and Assumption \ref{Bounded domain} are widely adopted to analyze stochastic optimization. Theorem \ref{Bounded domain} implies that, when optimizing a convex function, each iteration of the SGD is projected onto a constraint set $W$. This is a common assumption, as shown in paper \citep{shamir2013stochastic}.

\subsection{SGD, LAWA, and FWA}

\textbf{SGD.}
For the given training set $S=(z_1, z_2,..., z_n)$ and the target function $F$, the general update rule of the stochastic gradient descent (SGD) algorithm is formulated as 
\begin{equation}\label{SGD-rules}
w_{t} = w_{t-1} - \alpha_t \nabla_w F(w_{t-1},z_{i_t}),
\end{equation}
where $\alpha_t$ is the step size and $z_{i_t}$ is the sample chosen in iteration $t$. According to the gradient update rule, $w_{t}$ is represented as $w_{t} \!=\! w_0\! -\! \alpha_i\sum_{i=1}^{t}\nabla F(w_{i-1},z_{i})$, where $w_0$ is the initial point. 

In generalization, we discuss the following sampling method. Choosing $z_{i_t}$ with replacement is a standard way to train the model \citep{bottou2009curiously}. 
Two popular schemes are commonly used to select samples. One is to pick $i_t \sim Uniform\left\{1, \cdots , n \right\}$ at each step. The other is to choose a random permutation over $\left\{1, \cdots, n \right\}$ and cycle through the examples repeatedly in the order determined by the permutation. This setting is commonly explored in analyzing the stability  \citep{hardt2016train,xiao2022stability}. 

\textbf{LAWA.} Collecting model checkpoints once at the end of each epoch in a queue \citep{kaddour2022stop}, LAWA ’s solution at the end of epoch $E$ is
\begin{equation}\label{LAWA-rules}
w_{E}^{LAWA} = \frac{1}{k} \sum_{i=E-k+1}^{T} w_{i}.
\end{equation}

$k$ is a hyperparameter, generally a small constant in the practical setting. It has been empirically observed in \cite{kaddour2022stop} that when $k$ is relatively large, it leads to performance degradation. Therefore, we set a constant $k (1 \leq k \ll T)$ here. In the original version \citep{kaddour2022stop}, the LAWA collects one checkpoint from each of the last $k$ epochs. We may collect checkpoints every $m$ step under some training regime that we need \citep{sanyal2023early}, in which LAWA runs a moving window at a predetermined interval to collect $k$ latest checkpoints on the sequence of saved checkpoints $w_i$. 

\textbf{FWA.} 
We consider the general form to facilitate the theoretical analysis, averaging the last $k$ checkpoints over $T$ iterations. Then, FWA is defined as:
\begin{equation}\label{FWA-rules}
    \bar{w}_T^{k}=\frac{1}{k} \sum_{i=T-k+1}^{T} \rho_{i-(T-k)} w_{i},
\end{equation}
where $\rho_{i}$ is a weighting parameter that tunes the frequency of checkpoint saves when $\rho_{i}\in \left\{0,1\right\}$ and distinguishes the different averaging algorithms.  
\begin{remark}\label{rem-1}
Based on the Eq. ~\eqref{FWA-rules}, there are two special cases where FWA reduces to SGD when $k=1$ and $\rho_i = 1$, while SWA is obtained when $k=T$ and $\rho_{i} = 1$. 
Let $T=Ed$, where $E$ denotes the $E$ epochs and $d$ denotes $d$ iterations in each epoch. Then, LAWA is obtained when $1<k\ll T$ and $\rho_{i,1} = \cdots = \rho_{i,d-1} =0, \rho_{i,d}=1$. When $1<k\ll T$ and $\rho_{i}=1$, we derive the special FWA, i.e., "tail-averaging".
\end{remark}

We investigate FWA in the $1<k\ll T$ setting according to the above definition. The equivalent formulation of FWA's updating rules is written as:
\begin{equation}\label{lemma-FWA-update}
    \bar{w}^{k}_{T} = \bar{w}^{k}_{T-1} - \frac{1}{k}\sum_{i=T-k+1}^{T} \rho_{i-(T-k)}\alpha_i \nabla F(w_{i-1},z_i).
\end{equation}
We will use it a lot later in this analysis. The detailed derivation of Eq.~\eqref{lemma-FWA-update} is placed in the {\bf Appendix} \ref{pro-FWA-update}. Setting $\rho_i = 1$, Eq.~\eqref{lemma-FWA-update} degrades into $\bar{w}^{k}_{T} = \bar{w}^{k}_{T-1} - \frac{1}{k}\sum_{i=T-k+1}^{T} \alpha_i \nabla F(w_{i-1},z_i)$, and it can be simultaneously applied to both the tail-average algorithm and LAWA, with the only difference being the number of summation terms. 

\subsection{The Expansive Properties}
\begin{lemma}\label{lemma}
Assume that the function $F$ is $\beta$-smooth. Then, \\
{\bf (1). (non-expansive)} If $F$ is convex, for any $\alpha \leq \frac{2}{\beta}$, we have $\Vert w_{T+1}-w_{T+1}^{\prime} \Vert \leq \Vert w_{T}-w_{T}^{\prime}\Vert$; \\
{\bf (2). ($(1\!+\!\alpha\beta)$-expansive)} If $F$ is non-convex, for any $\alpha$, we have $\Vert w_{T+1}\!-\!w_{T+1}^{\prime} \Vert \!\leq\! (1\!+\!\alpha\beta)\Vert w_{T}\!-\!w_{T}^{\prime}\Vert$.
\end{lemma}

Lemma \ref{lemma} tells us that, in general, smoothness will imply that the gradient updates cannot be overly expansive. In addition, when the function is convex and the step size is sufficiently small, the gradient update becomes non-expansive. The proof of Lemma \ref{lemma} is deferred to {\bf Appendix} \ref{pro-lemma}. And more results can be found in several literature \citep{hardt2016train,xiao2022stability}. 

\subsection{Stability and Generalization Definition}

We can characterize the generalization error bound of an algorithm by controlling its uniform stability via this theorem. \citet{hardt2016train} link the \emph{uniform stability} of the learning algorithm to the expected generalization error bound and derive the generalization error bound for the SGD algorithm. The expected generalization error of a model $w = A_S$ trained by a certain randomized algorithm $A$ is defined as 
    \begin{equation}\label{D-gen}
\mathbb{E}_{S,A}\left[R_{S}\left[A_S\right]-R_\mathcal{D}\left[A_S\right]\right]. 
    \end{equation}
Next, we employ the following notion of \emph{uniform stability}.

\begin{definition}[$\epsilon$-Uniformly Stable]
A randomized algorithm $A$ is $\epsilon$-uniformly stable if for all data sets $S, S^{\prime} \in Z^{n}$ such that $S$ and $S^{\prime}$ differ in at most one example, we have
    \begin{equation}\label{E-stab}
     \mathop{sup}\limits_{z\in Z}\left\{\mathbb{E}_{A}\left[F(A_S;z)-F(A_{S^{\prime}};z)\right] \right\} \leq \epsilon.
    \end{equation}
\end{definition}
Here, the expectation is taken only over the internal randomness of $A$. 

\begin{theorem}{\rm (Generalization in Expectation, \cite[Theorem 2.2]{hardt2016train})}\label{Generalization in Expectation}
Let $A$ be $\epsilon$-uniformly stable. Then,
    \begin{equation}\label{E-gen}
     \vert\mathbb{E}_{S,A}\left[R_{S}\left[A_S\right]-R_\mathcal{D}\left[A_S\right]\right]\vert \leq \epsilon.
    \end{equation}
\end{theorem}

This theorem clearly states that if an algorithm has uniform stability, its generalization error is small. In other words, uniform stability implies \emph{generalization in expectation} \citep{hardt2016train}. Above proof is based on \cite[Lemma 7]{bousquet2002stability} and very similar to \cite[Lemma 11]{shalev2010learnability}.

\section{Generalization Bound under Convexity}\label{Generalization}

In this section, the generalization bounds will be constructed if we can control their uniform stability recursively using the properties of gradient updates in different scenarios. In contrast to SGD bounding only $\alpha_T \Vert \nabla F(w_{T-1},z_{T}) - \nabla F(w^{\prime}_{T-1},z_{T}) \Vert$, the difficulty of the analysis is that at step $T$ it is necessary to bound the cumulative gradient
\begin{equation}\label{cum-grad}
\frac{1}{k}\sum_{i=T-k+1}^{T} \rho_{i-(T-k)}\alpha_i \Vert \nabla F(w_{i-1},z_{i}) - \nabla F(w^{\prime}_{i-1},z_{i}) \Vert,
\end{equation}
which follows from the expression of $\bar{w}_{T}$ in Eq.~\eqref{lemma-FWA-update}. 

Another obstacle arises in the different learning rates. 
The constant and decaying learning rate result in significant differences in the mathematical formulation of accumulated gradients. This section will rigorously address these challenges through mathematical analysis. Due to limited space, we provide the detailed proof in the {\bf Appendix} \ref{pro-con}. 

\subsection{Convex Losses}
We investigate the generalization performance of FWA in light of Theorem \ref{Generalization in Expectation}, which indicates that the generalization bound is established through the stability bound. Then, assuming that the target function $F$ is convex and using the $L$-Lipschitz property of the target function, we have 
\begin{equation}\label{convex-basic}
\mathbb{E}\vert F(\bar{w}_{T}^k;z)-F(\bar{w}_T^{k\prime};z)\vert \leq L \mathbb{E} \Vert \bar{w}_T^k -\bar{w}^{k\prime}_T \Vert
 \end{equation}
for all $\bar{w}_T^k$ and $\bar{w}_T^{k\prime}$.
This implies that if the algorithm is stable, its generalization bound is immediately obtained because $\Vert \bar{w}_T^k -\bar{w}^{k\prime}_T \Vert$ is bounded.  

Next, we consider the stability bound in the case of sampling with replacement. The condition $\Vert \nabla F(u) \Vert \leq L$ from Assumption \ref{$L$-Lipschitz} and the non-expansive properties outlined in Lemma \ref{lemma} are utilized in this analysis.

\begin{theorem}\label{thm:stability-con-with}
Assume that the loss function $F(w;z)$ is convex, $L$-Lipschitz and $\beta$-smooth for all given $z\in\mathcal{Z}$ with sizes $n$. Suppose we run FWA with step sizes $\alpha_i \leq \frac{2}{\beta}$ for $T$ steps, where each step samples $z$ from $\mathcal{Z}$ uniformly with replacement. Then, FWA has uniform stability of
\begin{equation*}
  \epsilon_{gen} = \frac{2L^2}{nk} \left( \sum_{t=1}^{k}\sum_{i=1}^{t} \rho_i \alpha_i + \sum_{t=k+1}^{T}\sum_{i=t-k+1}^{t} \rho_{i-(t-k)} \alpha_i \right)
 \end{equation*}
where $\rho_i$ is a weighting parameter, see remark \ref{rem-1}  for a discussion of it.
\end{theorem}

\begin{remark}
To make Theorem \ref{thm:stability-con-with} valid, condition $\alpha_i \leq \frac{2}{\beta}$ must be satisfied. The rationale for this requirement is that when the function $F$ is convex, if the step size does not exceed $\frac{2}{\beta}$, the gradient updates will exhibit the non-expansiveness property, leading to a stable training process over time. In contrast, for convex problems, a larger learning rate can cause oscillations near the minima, preventing convergence.    
\end{remark}

\begin{corollary}\label{thm:stability-con-with-cor}
For any $\rho_i =1$, assume the same setting as Theorem \ref{thm:stability-con-with}.
Suppose we run FWA with constant step sizes $\alpha \leq \frac{2}{\beta}$ for $T$ steps, where each step samples $z$ from $\mathcal{Z}$ uniformly with replacement. Then, FWA has uniform stability of
\begin{equation}\label{stability-con-with-cor}
  \epsilon_{gen} = \mathbb{E}\vert F(\bar{w}_{T}^k;z)-F(\bar{w}_{T}^{k\prime};z)\vert \leq \frac{2\alpha L^2}{n} \left( T - \frac{k}{2} \right).
 \end{equation}
\end{corollary}

\begin{remark}
For FWA, when $k=1$, the stability bound $2\alpha L^2\left(T - \frac{k}{2} \right)/n $ in Corollary \ref{thm:stability-con-with-cor} simplifies to $2\alpha L^2 T/n$ for SGD, as shown in \cite{hardt2016train}. This implies that the larger the number of averages $k$ in FWA under convex assumptions, the smaller the generalization bound. Here, we obtain a more general result, and for a fixed $k$, the bound of FWA shows a linear improvement compared to SGD. As illustrated in \Cref{fig:con-gen-para-fix,fig:con-gen-acc-fix}, this phenomenon is observed in our experiments.
\end{remark}

\begin{remark}
Based on Theorem \ref{thm:stability-con-with}, we can derive the generalization bounds under different learning rate settings. Let $\rho_i = 1$. When $\alpha_i = \alpha = \frac{c}{T}$, the generalization bound is $\frac{2cL^2}{n}(1-\frac{k}{2T})$; when $\alpha_i = \frac{c}{t}$, the generalization bound is $\frac{2cL^2}{n}(1+\log(T))$. A decaying learning rate corresponds to a larger bound, which can be attributed to the larger step size. We compare the stability and generalization results w/ and w/o learning rate decay in \Cref{fig:con-gen}, which are consistent with our analysis.
\end{remark}

\begin{corollary}
For any $\rho_i =1$ and $k=T$, assume the same setting as Theorem \ref{thm:stability-con-with}.
Suppose we run SWA with constant step sizes $\alpha \leq \frac{2}{\beta}$ for $T$ steps, where each step samples $z$ from $\mathcal{Z}$ uniformly with replacement. Then, SWA has uniform stability of
\begin{equation}
  \epsilon_{gen} = \mathbb{E}\vert F(\bar{w}_{T}^k;z)-F(\bar{w}_{T}^{k\prime};z)\vert \leq \frac{\alpha T L^2}{n}.
 \end{equation}
\end{corollary}

\begin{remark}
When $k=T$, the bound reduces to $\alpha L^2 T/n$ for SWA, as demonstrated in \cite{hardt2016train,xiao2022stability,pmlr-v235-wang24bl}. As demonstrated in \Cref{fig:con-gen}, the generalization ability of FWA is weaker than SWA in the case of convex, which can be easily explained from the landscape perspective. SWA makes full use of the information of each checkpoint during the training process, and it is easy to find the global optimum in the solution space. However, FWA emphasizes the exploration of the local properties of the solution space by averaging the checkpoints collected in the tail phase during training. Especially when $k$ is small, it is more likely to find local minima.
\end{remark}

When the function is convex, the effect of the learning rate on the generalization bound is intuitive, as it mainly depends on how to choose $T$ while not affecting the mathematical form of the generalization bound. However, for non-convex cases, it becomes more complicated since the bound is affected by the $(1+\alpha \beta)$-expansive property and accumulated gradients. Consequently, our discussion will focus on how different learning rate schedules affect the generalization bound in the non-convex case.

\subsection{Non-Convex Losses}
This section provides the generalization bound based on the non-convex assumption and $(1+\alpha\beta)$-expansive properties in Lemma \ref{lemma}, making the proof difficult. The goal is achieved through building its stability bound, and based on Theorem \ref{Generalization in Expectation}, we apply the $L$-Lipschitz on $F(\cdot;z)$ for fixed example $z \in \mathcal{Z}$ to get 
\begin{equation}\label{nonconvex-basic}
\mathbb{E}\vert F(\bar{w}_{T}^k;z)-F(\bar{w}_{T}^{k\prime};z)\vert \leq \frac{t_0}{n} + L \mathbb{E}\left[\Vert \bar{w}_T^k -\bar{w}^{k\prime}_T \Vert \vert \Vert \bar{w}_{t_0}^k-\bar{w}_{t_0}^{k\prime}\Vert=0\right],
\end{equation}
where 
$t_0 \in [0,n]$. The ideas of the non-convex case are motivated by the arguments in \cite{hardt2016train}, where the objective is divided into two parts, whether the different samples are selected before step $t_0$, with its proof in Appendix \ref{proof-noncon-basic}. Next, we will derive the stability bound under the non-convex setting based on Eq.~\eqref{nonconvex-basic}.

\begin{lemma}\label{Lemma_noncon}
Assume the target function $F$ is $\beta$-smooth and $non$-convex. $w_T$ and $w_T^{\prime}$ denote the output after $T$ steps of SGD on datasets $S$ and $S^{\prime}$ where differing in only a single example. \\
For constant step sizes $\alpha_t = \frac{c}{T}$, we have
  \begin{equation}
  \begin{aligned}
   \Vert w^{\prime}_{T}& - w_{T} \Vert \leq &e^\frac{c\beta k}{T}\bar{\delta}_{T},
  \end{aligned}
 \end{equation}
For decaying step sizes $\alpha_t = \frac{c}{t}$, we have
  \begin{equation}
  \begin{aligned}
   \Vert w^{\prime}_{T}& - w_{T} \Vert \leq &e^\frac{c\beta k}{T-k}\bar{\delta}_{T},
  \end{aligned}
 \end{equation}
\end{lemma}
where $\bar{\delta}_{T}=\frac{1}{k}\sum_{i=T-k+1}^{T}\Vert w_{i} - w_{i}^{\prime} \Vert$, $c>0$ is a constant and $1\leq k \ll T$ denotes the number of averaging. Lemma \ref{Lemma_noncon} transforms the parameter difference at the T-th step of the accumulated gradient into the average parameter difference, which helps us establish the recursive relation for the averaging algorithm. This is a key aspect that highlights the technical feature of model averaging. The proof of Lemma \ref{Lemma_noncon} primarily relies on the $(1+\alpha\beta)$-extension property outlined in Lemma \ref{lemma}. The detailed proof can be found in the Appendix \ref{proof-Lemma_noncon}.

\begin{theorem}\label{thm:stability-non-with}
For any $\rho_i = 1$, assume that the loss function $F(w;z)\in [0,1]$ is non-convex, $L$-Lipschitz and $\beta$-smooth for all given $z\in\mathcal{Z}$ with sizes $n$. Suppose we run FWA with constant step sizes $\alpha \leq \frac{c}{T}$ for $T$ steps, where each step samples $z$ from $\mathcal{Z}$ uniformly with replacement. Then, FWA has uniform stability of
\begin{equation}\label{result-5.3}
  \epsilon_{gen} = \mathbb{E}\vert F(\bar{w}_{T}^k;z)-F(\bar{w}_{T}^{k\prime};z)\vert \leq \mathcal{O} \left(\frac{T^{\frac{c\beta}{c\beta+k}}}{n}\right).
 \end{equation}
\end{theorem}
We leave the details of this proof in Appendix \ref{proof-thm-non-with}.

\begin{remark}
The generalization bound $\mathcal{O}(T^{\frac{c\beta}{c\beta+k}})$ for FWA in Theorem \ref{thm:stability-non-with} shows an improvement over the existing bound $\mathcal{O}(T^{\frac{c\beta}{c\beta+1}})$ for SGD, as reported in \citep{hardt2016train}. This enhancement suggests that FWA can transform the $(1+\alpha\beta)$-expansive of SGD into $(1+\frac{\alpha\beta}{k})$-expansive, which implies that the averaged model has altered the expansive properties of the stochastic gradient method. The accumulation of $T$-step iterations builds the generalization bound, so each smaller expansion step allows for larger generalization gains. Therefore, this improvement and theoretical explanation are fundamental.
\end{remark}

In the non-convex setting, the generalization upper bound for FWA is influenced by the expansion properties and accumulated gradients. Larger accumulated gradients lead to a larger generalization bound. Notably, the learning rate during the averaging stage is crucial in affecting the accumulated gradients. Next, we will present the generalization bound under a decaying learning rate.

\begin{theorem}\label{thm:stability-non-with-decay}
Assume the same setting as Theorem \ref{thm:stability-non-with}. Suppose we run FWA with decaying step sizes $\alpha \leq \frac{c}{t}$ for $T$ steps, where each step samples $z$ from $\mathcal{Z}$ uniformly with replacement. FWA has uniform stability of
\begin{equation}
  \mathbb{E}\vert F(\bar{w}_{T}^k;z)-F(\bar{w}_{T}^{k\prime};z)\vert \leq \mathcal{O}\left(\frac{T^{\frac{kc\beta+c^2 \beta^2}{2kc\beta+c^2 \beta^2 +k^2 (1-c\beta)}}}{n} \right).
 \end{equation}
\end{theorem}
When ${\frac{kc\beta+c^2 \beta^2}{2kc\beta+c^2 \beta^2 +k^2 (1-c\beta)}}<\frac{c\beta}{1+c\beta}$, i.e., $k^2-k>\frac{c\beta}{1-c\beta}$ is satisfied, the upper bound of FWA is smaller than SGD. However, under the condition $c\beta \in (0,1)$ and $k \in (1,\frac{1+\sqrt{1+4c\beta}}{2})$, numerical verification shows that no $c\beta$ and $k$ satisfy condition $k^2-k>\frac{c\beta}{1-c\beta}$. This indicates that the generalization upper bound of FWA is greater than that of SGD. The primary reason for this outcome lies in the decaying learning rate in FWA, which results in greater accumulation gradients, thereby leading to a larger generalization bound. The detailed proof process is provided in Appendix \ref{proof-thm-non-with-decay}.  

\begin{figure}[t]
    \centering
    \begin{subfigure}{0.45\linewidth}
    \includegraphics[width=\linewidth]{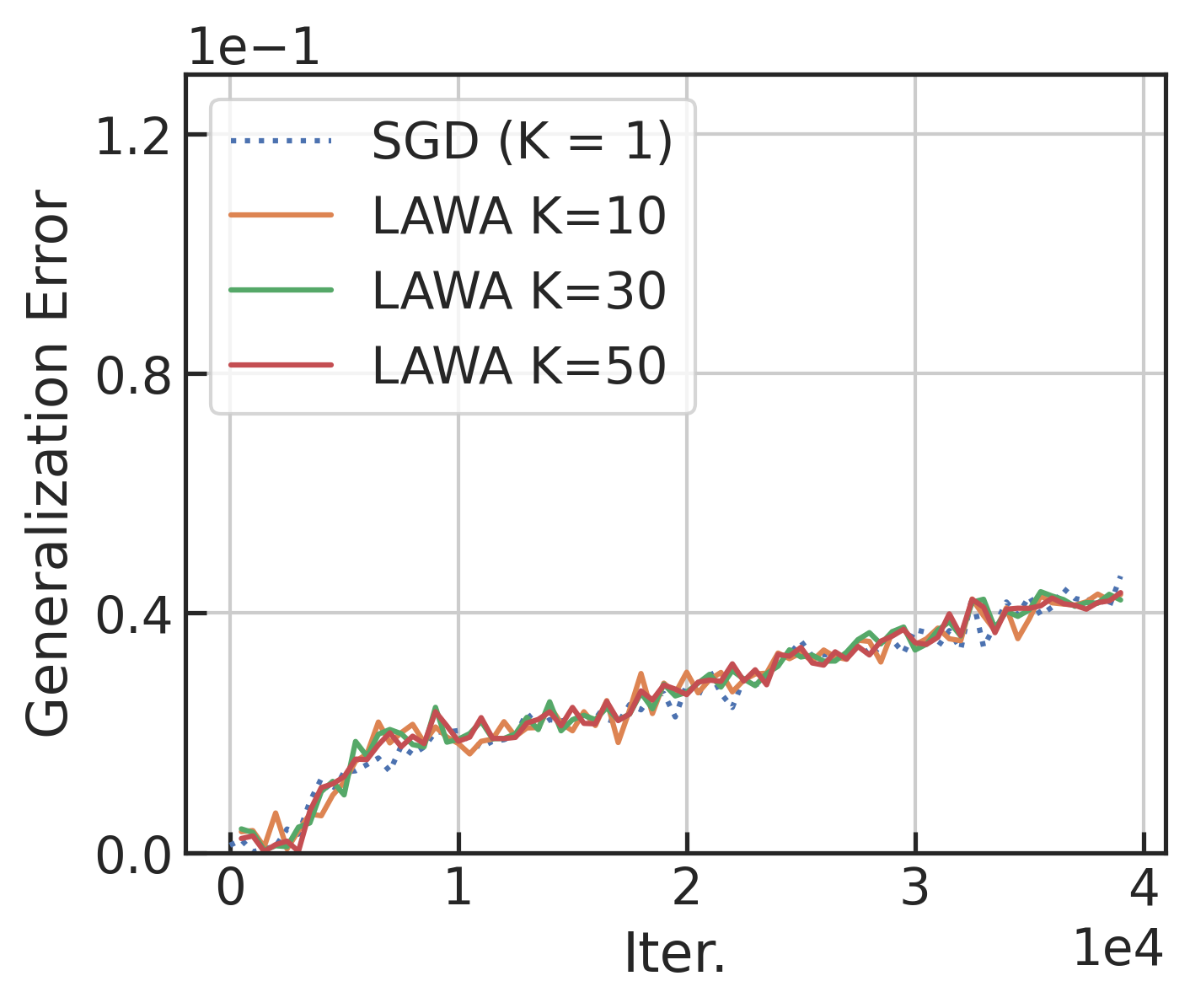}
    \caption{Generalization w/o learning rate decay}
    \end{subfigure}
    \hfill
    \begin{subfigure}{0.45\linewidth}
    \includegraphics[width=\linewidth]{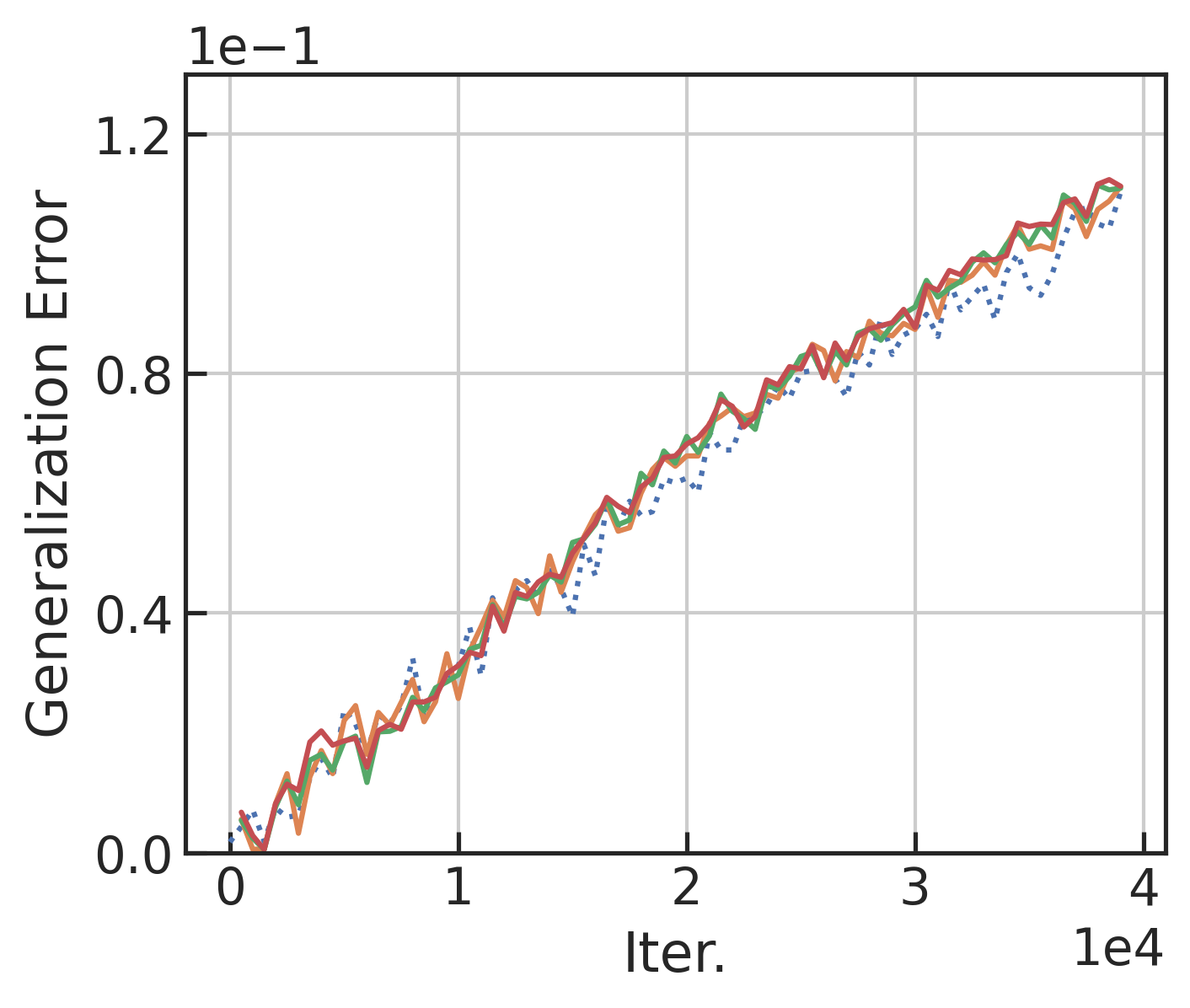}
    \caption{Generalization w/ learning rate decay}
    \end{subfigure}
    \caption{Generation results of the CIFAR10 dataset.}\label{fig:con-gen-cifar-2}
\end{figure}

\begin{remark}
The result of Theorem \ref{thm:stability-non-with-decay} indicates that, in the case of non-convex optimization with a decaying learning rate, the generalization bound of FWA is not always smaller than that of SGD. In other words, averaging a finite number of points at the tail end of the training trajectory does not consistently improve generalization. As shown in \Cref{fig:con-gen-cifar-2}, under the setting of decaying learning rate, the generalization error of FWA could be larger than SGD, which is not observed with constant learning rate under the same averaging steps. Viewed from the landscape, it is easy to provide examples where selecting points from one side of the minima in the training trajectory containing only a local minimum and averaging them leads to a deviation from the local minimum. Based on this, a larger learning rate means that the averaging stage incorporates updated information from earlier stages, which hinders the model to reach a flat minimum.
\end{remark}




\begin{remark}
Compared to Theorem \ref{thm:stability-non-with}, Theorem \ref{thm:stability-non-with-decay} with decaying learning rate exhibits a significantly larger generalization bound. Although both theorems satisfy the same $(1+\frac{\alpha\beta}{k})$-expansion property, the order of the accumulated gradient's bound plays a more critical role in this context. Specifically, for decaying learning rate in Theorem \ref{thm:stability-non-with-decay}, the order of the accumulated gradient is $\mathcal{O}\left(\left(1/t\right)^{1-\frac{kc\beta}{k+c\beta}}\right)$, while in Theorem \ref{thm:stability-non-with}, the order is $\mathcal{O}\left(\frac{1}{t}\right)$ (See \Cref{proof-thm-non-with-decay,proof-thm-non-with} for details). The larger order of accumulated gradient directly leads to a larger minimum value of $t_0$, which in turn results in a larger generalization bound.
\end{remark}

\begin{remark}
Under the $non$-convex setting, the assumption that $F(w;z) \in [0,1]$ in Theorem \ref{thm:stability-non-with-decay} and Theorem \ref{thm:stability-non-with} is made for simplicity of presentation. Removing this condition does not change the final results, as it only scales them by a constant factor. This is a relatively common assumption discussed in \cite{hardt2016train,xiao2022stability}. 
\end{remark}

\begin{figure}[t]
    \centering
    \begin{subfigure}{0.45\linewidth}
    \includegraphics[width=\linewidth]{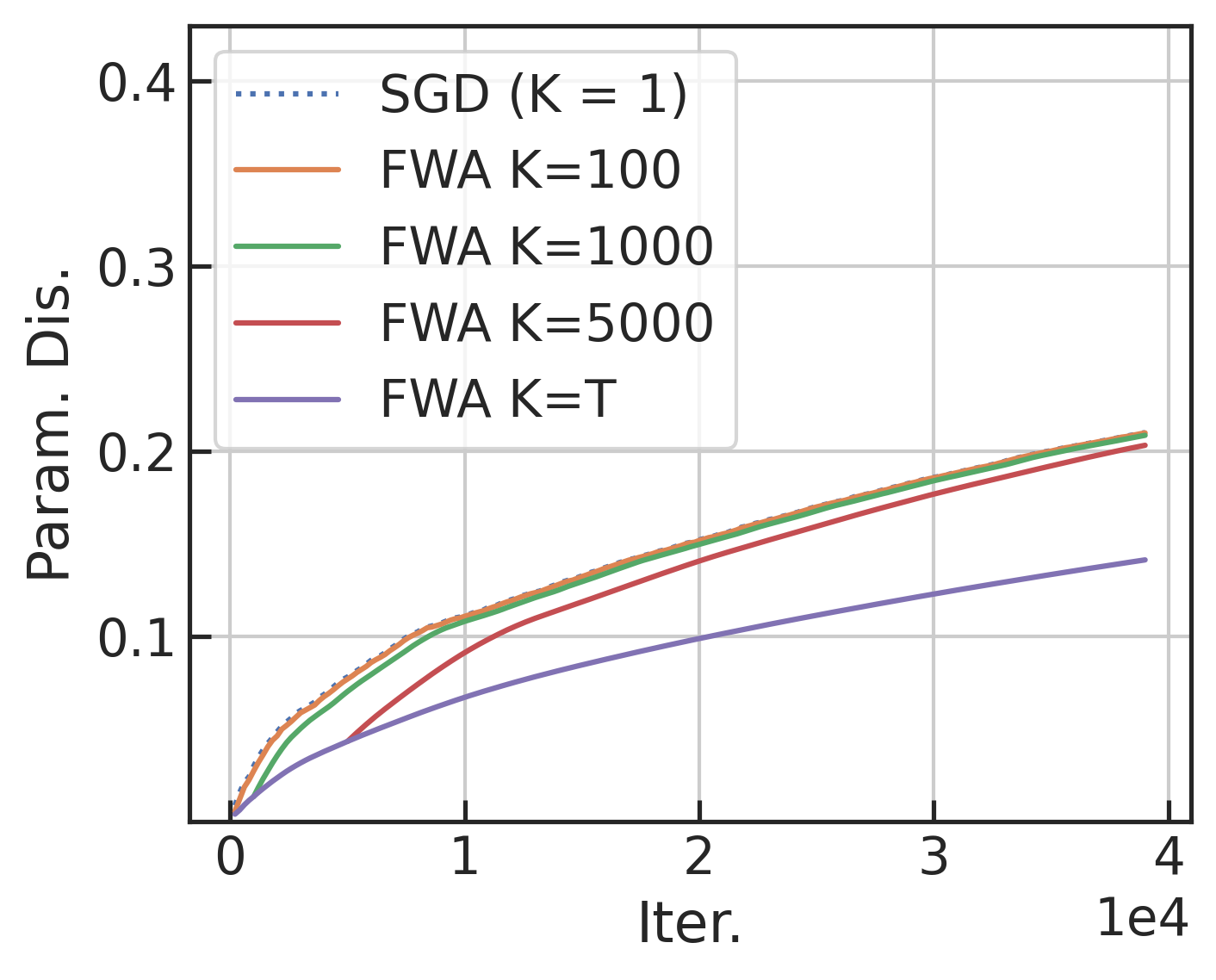}
    \caption{Stability w/o learning rate decay}\label{fig:con-gen-para-fix-cifar}
    \end{subfigure}
    \hfill
    \begin{subfigure}{0.45\linewidth}
    \includegraphics[width=\linewidth]{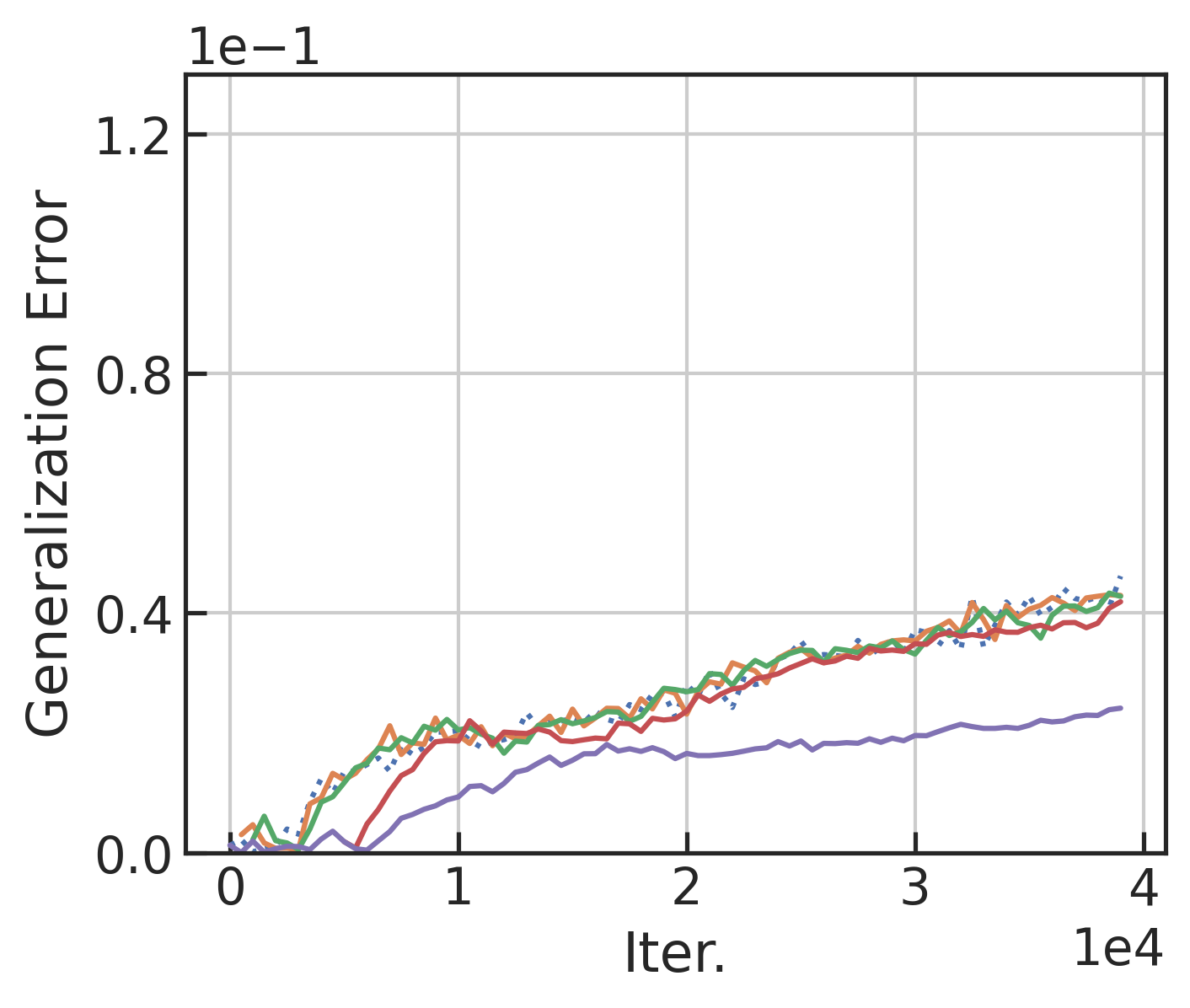}
    \caption{Generalization w/o learning rate decay}\label{fig:con-gen-acc-fix-cifar}
    \end{subfigure}
    \caption{Stability and generation results of the CIFAR10 dataset with constant learning rate.}\label{fig:con-gen-cifar}
\end{figure}

\textbf{Early stopping} has been demonstrated to improve the generalization ability of models. This phenomenon can be explained through the generalization bound based on stability, which depends on the number of training steps $T$ and decreases as $T$ decreases.
Similar to SGD,
the generalization bound of FWA also depends on the magnitude of the cumulative gradient. For instance, under constant learning rate settings, when the function $F$ is convex, the cumulative gradient is $\frac{2L}{nk}\sum_{i=T-k+1}^{T}\alpha_i$, and the generalization bound is $\frac{2\alpha L^2}{n}(T-\frac{k}{2})$ (see Theorem \ref{thm:stability-con-with-cor}); when the function is non-convex, the cumulative gradient is $\mathcal{O}(\frac{1}{T})$, and the generalization bound is $\mathcal{O}(T^{\frac{c\beta}{c\beta+k}}/n)$ (see Theorem \ref{thm:stability-non-with}). An increase in the number of training steps $T$ directly leads to an increase in the magnitude of the cumulative gradient with respect to $T$, which in turn increases the generalization bound. This explains why early stopping results in a smaller generalization bound. Our experiments, as shown in \Cref{fig:con-gen-cifar}, also validate this observation.

\section{Convergence Analysis}\label{sec:convergence}

In this section, we explore the convergence of the finite averaging schemes based on the last iterates, which return some weighted combination of the last $k$ iterates $w_{T-k+1}, \cdots, w_T$. We focus here mainly on the convex case to compare with SGD and the already known simple averaging of all iterations, as shown in Table \ref{sample-table2}.


\begin{proposition}\label{thm:convergence1}
 Under the assumptions \ref{Convex function}, considering the FWA with step sizes $\alpha_t$ and weighted sequences $\left\{\rho_i\right\}$. Then, for any $T>1$ and $1\leq k\leq \frac{T}{2}$, it holds that:
\begin{equation}\label{result-4.1}
\begin{aligned}
  \mathbb{E} [F(\bar{w}_T^k)- &F(w))] \leq  \sum_{t=T-k+2}^{T} \mathbb{E}\Vert w_{t} - w \Vert^2 \left(\frac{\rho_{t-(T-k)}}{2k\alpha_t}-\frac{\rho_{t-(T-k+1)}}{2k\alpha_{t-1}} \right) \\ & + \frac{\rho_1 \mathbb{E}\Vert w_{T-k+1} - w \Vert^2}{2k\alpha_{T-k+1}} +\frac{G^2}{2k} \sum_{t=T-k+1}^{T} \rho_{t-(T-k)}\alpha_t .    
\end{aligned}
 \end{equation}   
\end{proposition}
 
Proposition \ref{thm:convergence1} focuses on the convex and non-smooth assumptions and includes a weighting parameter $\rho_i$ and a learning rate $\alpha_t$ in its results. Precisely, different values of $\rho_i$ determine distinct algorithmic forms, such as SGD, LAWA, and tail-averaging algorithms. In the following parts, we will discuss the convergence outcomes under varying values of these parameters. The detailed proof is placed in the \textbf{Appendix} \ref{proof-thm-convergence}.

We first establish the convergence bounds for the FWA, also known as the "tail-averaging" algorithm, when $\rho_i = 1$. The detailed proof is placed in the \textbf{Appendix} \ref{proof-thm-convergence-cor}.

\begin{theorem}\label{thm:convergence-cor}
For any $\rho_i = 1$ and under the assumptions \ref{Convex function}, \ref{Bounded variance} and \ref{Bounded domain}, considering the FWA with step sizes $\alpha_t = \frac{c}{\sqrt{t}}$, where $c>0$. Then, for any $T>1$ and $1\leq k\leq \frac{T}{2}$, it holds that: 
\begin{equation}\label{result-conv-cor}
  \mathbb{E}[ F(\bar{w}_{T}^k) - F(w^{\star})] \leq \frac{2+\log\left(\frac{T}{2k}\right)}{\sqrt{T}}\left(\frac{D^2}{c}+2cG^2\right),
 \end{equation}
where $D, G$ are some constants.
\end{theorem}

\begin{theorem}\label{thm:convergence-sgd}
For any $\rho_i = 1$, $k=1$ and under the assumptions \ref{Convex function}, \ref{Bounded variance} and \ref{Bounded domain}, considering the SGD with step sizes $\alpha_t = \frac{c}{\sqrt{t}}$, where $c>0$. Then, for any $T>1$, it holds that: 
\begin{equation}\label{result-conv-sgd}
  \mathbb{E}[ F(w_{T}) - F(w^{\star})] \leq \frac{2+\log{T}}{\sqrt{T}}\left(\frac{D^2}{c}+2cG^2\right),
 \end{equation}
where $D, G$ are some constants.
\end{theorem}

When $\rho_i=1$ and $k=1$, this result covers the last iterate convergence of SGD as a special case, which generalizes convergence to $\mathcal{O}(\log T/\sqrt{T})$. A similar theoretical result on SGD is given in \cite{shamir2013stochastic}. The detailed proof is placed in the \textbf{Appendix} \ref{proof-thm-convergence-sgd}.

\begin{figure*}[t]
    \centering
    \begin{subfigure}{0.45\linewidth}
    \includegraphics[width=\linewidth]{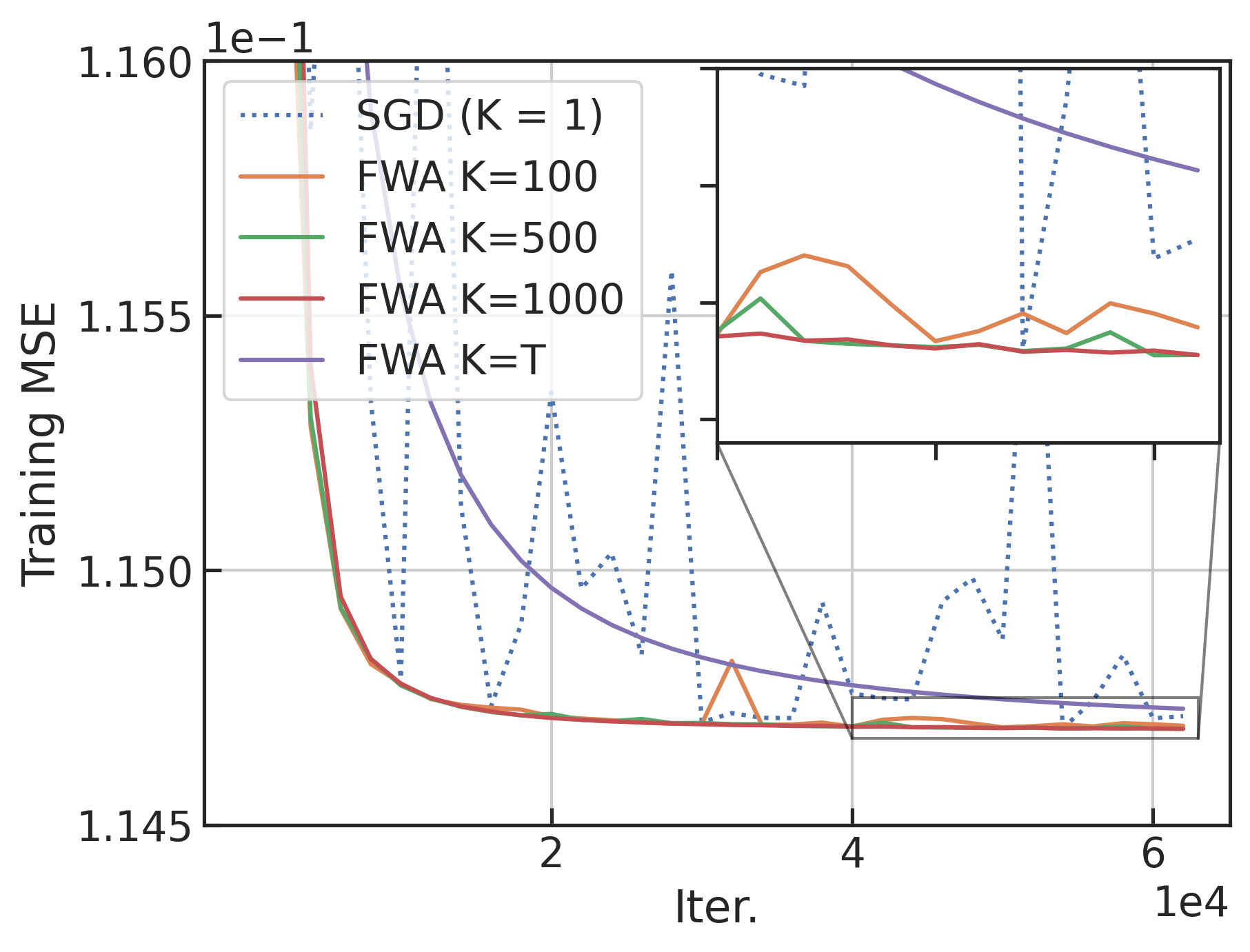}
    \caption{Training error w/o learning rate decay}\label{fig:convex-conergence-fixed-lr}
    \end{subfigure}
    \hfill
    \begin{subfigure}{0.45\linewidth}
    \includegraphics[width=\linewidth]{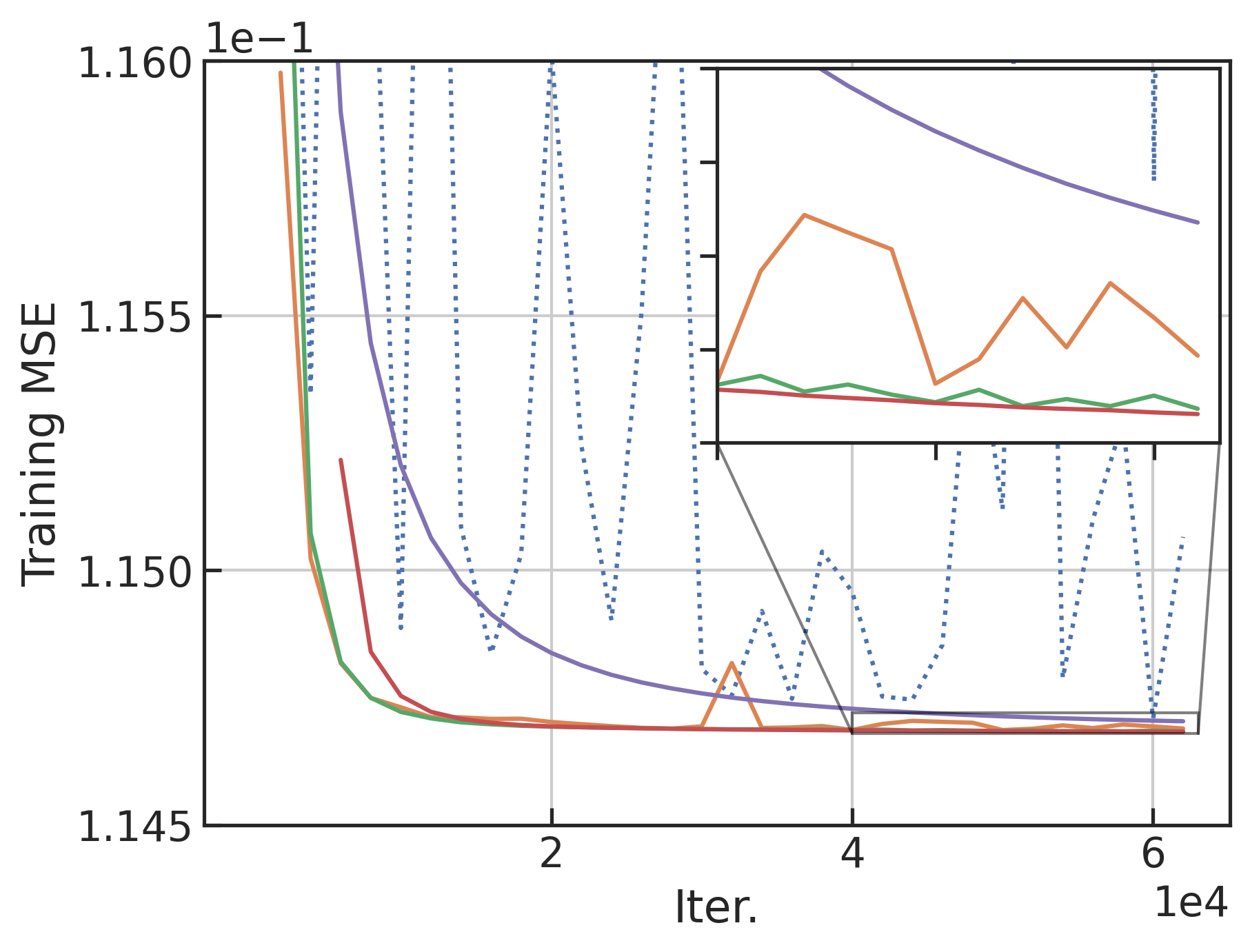}
    \caption{Training error w/ learning rate decay}\label{fig:convex-conergence-decaying-lr}
    \end{subfigure}
    \caption{Convergence results of the convex problem.}
    \label{fig:convex-convergence}
\end{figure*}

\begin{remark}
Theorem \ref{thm:convergence-cor} suggests that under convex assumptions, FWA has faster convergence compared to SGD in Theorem \ref{thm:convergence-sgd}, which implies averaging checkpoints during the final stages of training helps accelerate convergence. Moreover, for $k>1 (k\in(1, T/2))$, the larger the final averaging iterations, the faster the convergence. Our experiments in \Cref{fig:convex-convergence} validate these results.
\end{remark}

\begin{remark}
In Theorem \ref{thm:convergence-cor}, $k$ is constrained within the interval $\left[1, T/2 \right]$, which implies that when $k$ exceeds this range, an increase in average points does not necessarily lead to faster convergence. In practice, we observe that as $k$ increases, the convergence rate initially accelerates but then decelerates. This phenomenon occurs because averaging the most recent or tail-end points helps accelerate convergence when the training process oscillates near a local minimum. However, including too many points, such as those from earlier stages of training, may lead to deviation from the local minimum. Moreover, the optimal value of $k$ is often not very large, so this paper has consistently emphasized the importance of parameter $k\ll T$. This phenomenon was also observed in our experiments, as shown in \Cref{fig:convex-convergence}, and similar experimental results are validated in paper \cite{kaddour2022stop}.
\end{remark}

\begin{theorem}\label{thm:convergence-LAWA}
The training steps $T$ are sliced into $E$ epochs, each containing $d$ steps, i.e., $T=Ed$. For the last $k$ epochs, setting $\rho_{i,d} = d$ and $\rho_{i,1} = \cdots = \rho_{i,d-1} = 0$, where $i \in [E-k+1, E]$. Under the assumptions \ref{Convex function}, \ref{Bounded variance} and \ref{Bounded domain}, considering the LAWA with step sizes $\alpha_t = \frac{c}{\sqrt{t}}$, where $c>0$. Then, for any $E>1$, $d>1$, and $1\leq k\leq \frac{Ed}{2}$, it holds that: 
\begin{equation}\label{result-conv-LAWA}
  \mathbb{E}[ F(\bar{w}_{T}^{k}) - F(w^{\star})] \leq \frac{2d+d\log{\left(\frac{T}{2kd}\right)}}{\sqrt{T}}\left(\frac{D^2}{c}+2cG^2\right),
 \end{equation}
where $D, G$ are some constants.
\end{theorem}

\begin{remark}
Theorem \ref{thm:convergence-LAWA} suggests that the convergence bound of the LAWA depends not only on the iterations $T$ and the averaging number $k$ but also on the interval length $d$. As $d$ increases, the convergence bound $B$ decreases, implying that averaging checkpoints from earlier iteration in the end training stage may not benefit model convergence, which aligns with our intuition. Specifically, when $d=1$, this bound reduces to the FWA; when both $d=1$ and $k=1$, it further reduces to the SGD. The detailed proof is placed in the \textbf{Appendix} \ref{proof-thm-convergence-LAWA}. In \Cref{fig:convex-convergence-multi-steps}, our experiments show that convergence is slower as $d$ increases.
\end{remark}

Our results try to make them as applicable as possible to practical problems. In particular, smoothness often does not hold in deep learning. For example, the support vector machine optimization problem with the standard non-smooth hinge loss. Although we do not discuss strongly convex problems here, our analytical approach is equally applicable. Using a similar technique, \citet{shamir2013stochastic} obtain an individual iterate bound for SGD in the case of strongly convex and non-smooth, and \citet{zhang2004solving} solves the large-scale linear prediction problem with constant learning rate.

\begin{figure}[t]
    \centering
    \begin{subfigure}{0.45\linewidth}
    \includegraphics[width=\linewidth]{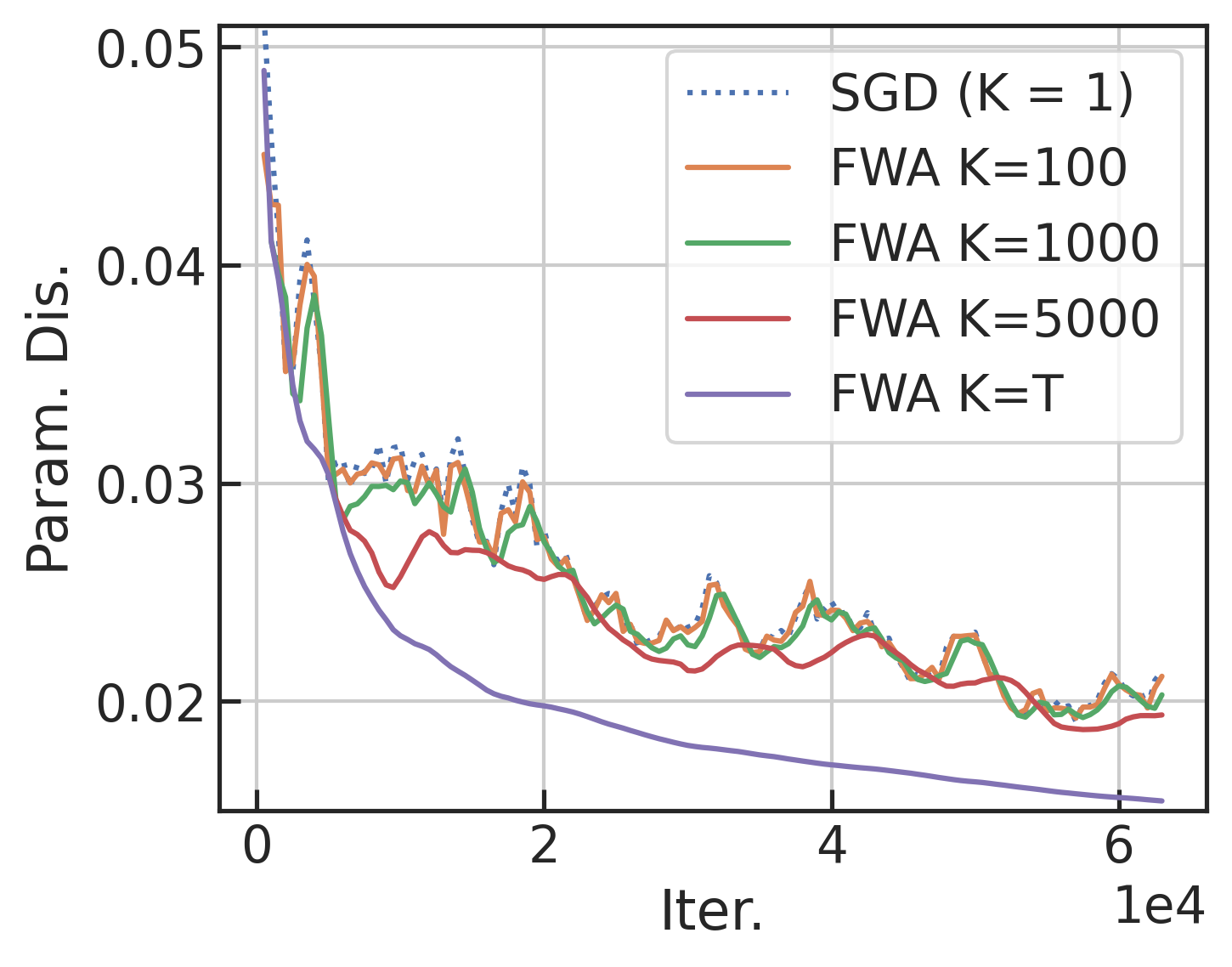}
    \caption{Stability w/o learning rate decay}\label{fig:con-gen-para-fix}
    \end{subfigure}
    \hfill
    \begin{subfigure}{0.45\linewidth}
    \includegraphics[width=\linewidth]{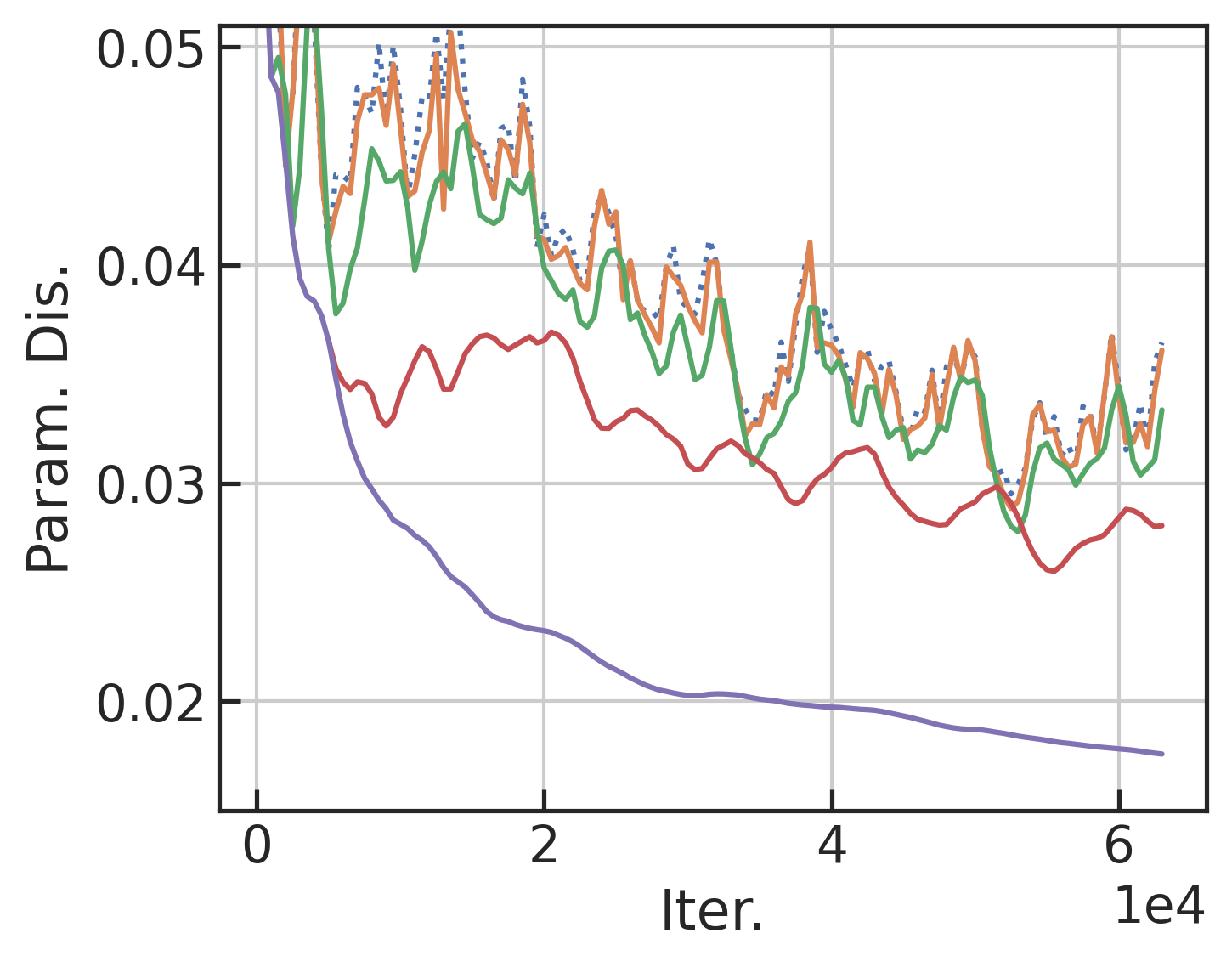}
    \caption{Stability w/ learning rate decay}
    \end{subfigure}
    \begin{subfigure}{0.45\linewidth}
    \includegraphics[width=\linewidth]{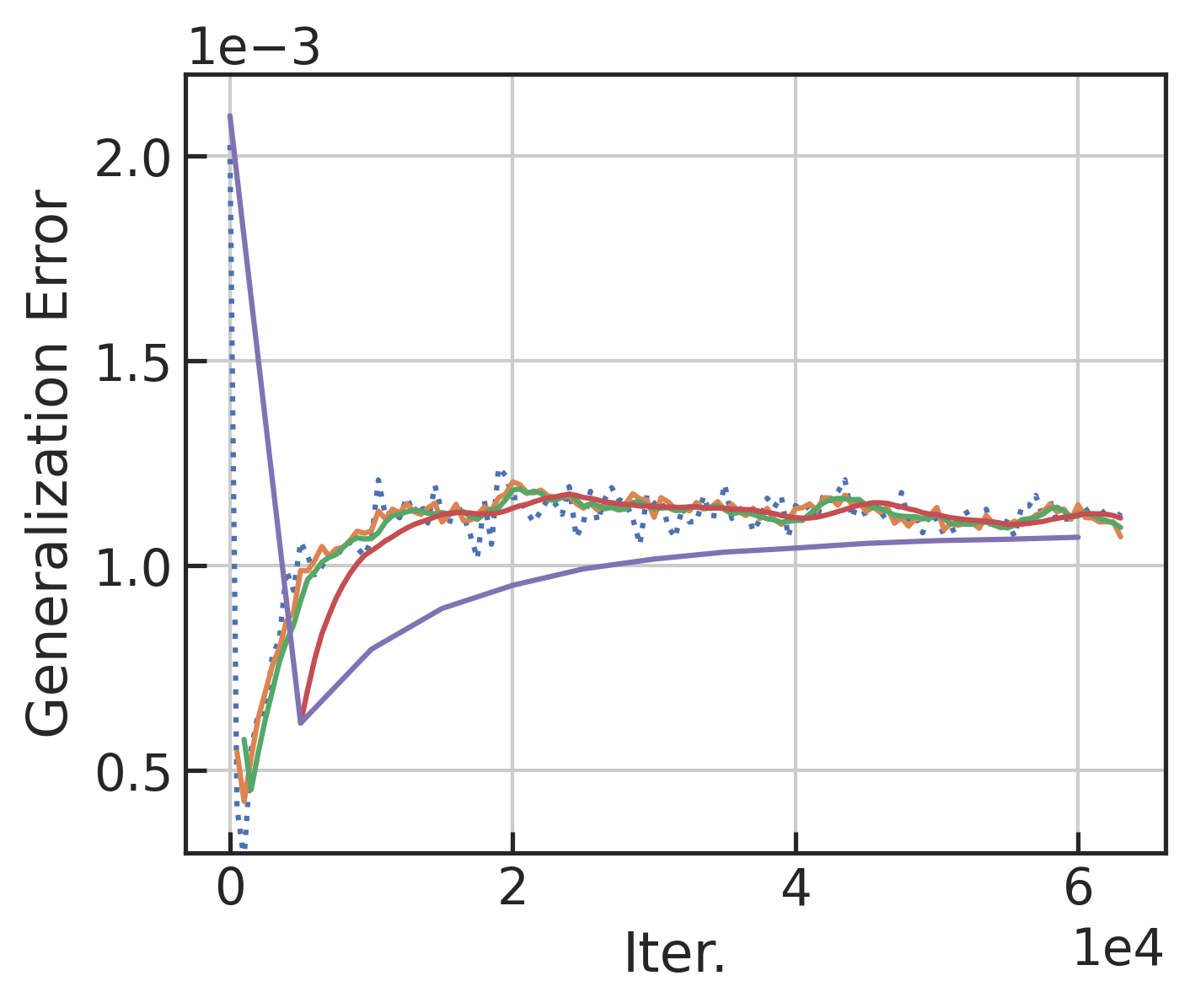}
    \caption{Generalization w/o learning rate decay}\label{fig:con-gen-acc-fix}
    \end{subfigure}
    \hfill
    \begin{subfigure}{0.45\linewidth}
    \includegraphics[width=\linewidth]{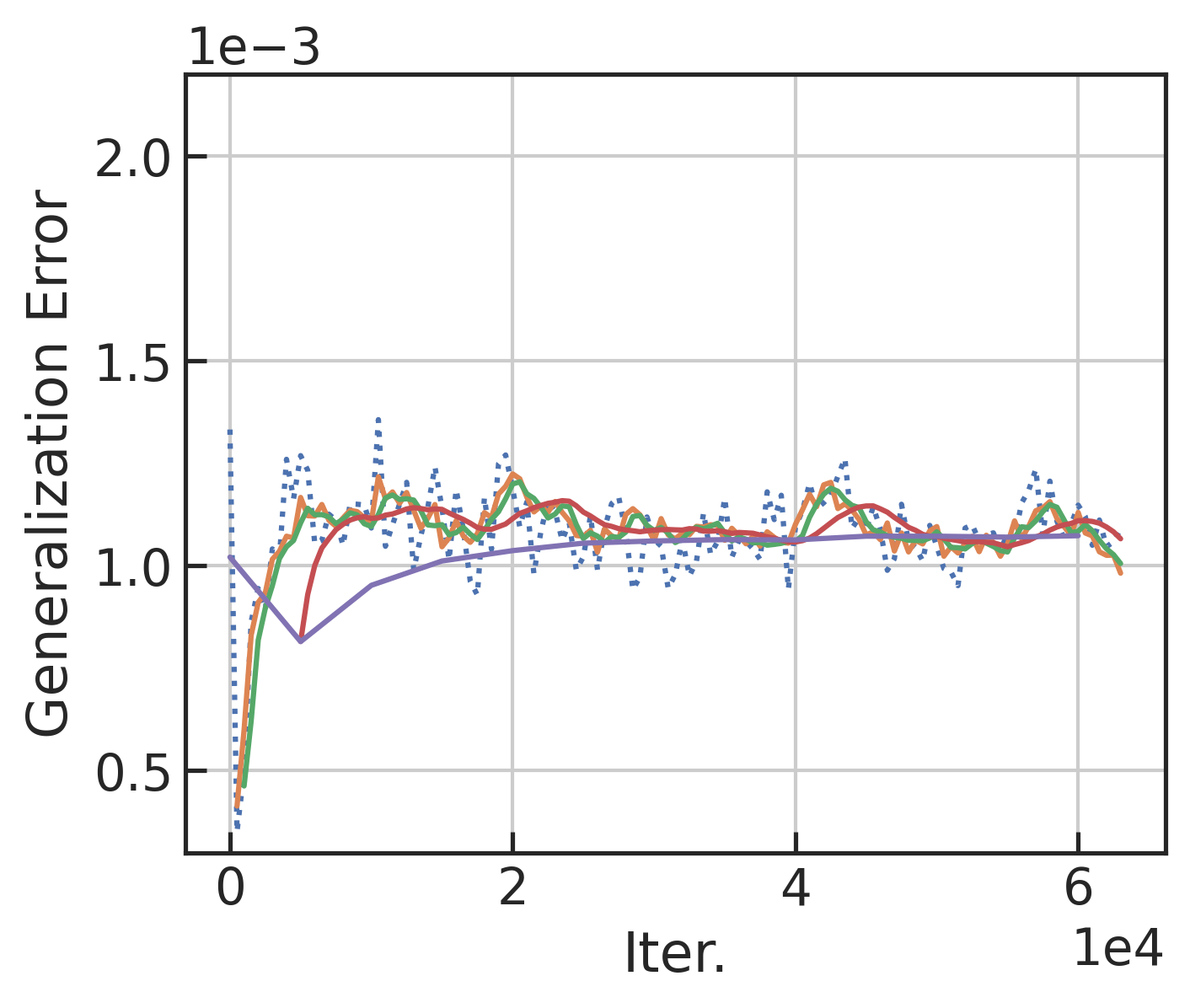}
    \caption{Generalization w/ learning rate decay}
    \end{subfigure}
    \caption{Stability and generation results of the Adult dataset.}\label{fig:con-gen}
\end{figure}

\section{Experimental Evaluation}
Below, we conduct several experiments to verify our derived optimization and generalization bound, respectively.
All experiments are conducted using the NVIDIA RTX A5000 with 24GB GPU memory, CUDA v11.3 and cuDNN v8.3.2 in PyTorch v1.12.1.

\paragraph{Setting and datasets.}

We consider convex and non-convex problems.
For the convex problem, we choose the Adult dataset from the UCI repository \citep{misc_adult_2}, which records adult incomes given 99 attributes. The input features are normalized using $l_1$ norm. A linear regression model is optimized by minimizing the mean squared error (MSE).
We adopt SGD with batch size 60.
For the constant learning rate case, the learning rate is set to $0.2$. For the decaying learning rate case, the learning rate starts from $0.4$, and decays four times uniformaly during the training, e.g., $\alpha_t = \frac{0.4}{\sqrt{t}}, t=1, 2, 3, 4$, so the the learning rate is converged to $0.2$.
For both cases, the model is optimized for 100 epochs.
For the FWA method, the averaging step $k$ is chosen from $\{1, 100, 1000, 5000, T\}$. Specifically, for $k = 1$, FWA reduces to SGD, while for $k = T$, all the historical models are averaged. 

Then we consider the non-convex problem of training LeNet \cite{lecun1998gradient} on the CIFAR10 dataset \cite{krizhevsky2009learning}.
Similarly, we adopt SGD with batch size 60.
For the constant learning rate case, the learning rate is set to $0.1$. For the decaying learning rate case, the learning rate starts from $0.2$, and then decays to $0.1$ as in the convex setting.
The model is also optimized for 100 epochs. Additionally, the averaging step $k$ is chosen from $\{1, 10, 30, 50\}$.

\paragraph{Evaluation.}

For the stability analysis, we follow the setting of \cite{hardt2016train}.
In specific, the stability is measured by the parameter distances trained on two different datasets and the generalization error.
Smaller parameter differences and generalization error indicate stronger stability.
To construct two different datasets,
we randomly remove one sample from the training set to construct dataset $S$. Then, another dataset $S'$ is constructed by replacing one random data point in $S$ with the deleted one. We train the model on these two datasets with the same initialization and settings. The parameter distance is defined as the Euclidean distance between parameters of the two models, \emph{e.g.},
$\sqrt{\Vert w-w^{\prime}\Vert^2/(\Vert w \Vert^2 + \Vert w^{\prime}\Vert^2)}$, where $w$ and $w^{\prime}$ denote all the parameters of models trained on $S$ and $S'$ respectively.
The generalization error is the absolute value of the difference between the training and test errors.

\begin{figure*}[t]
    \centering
    \begin{subfigure}{0.45\linewidth}
    \includegraphics[width=\linewidth]{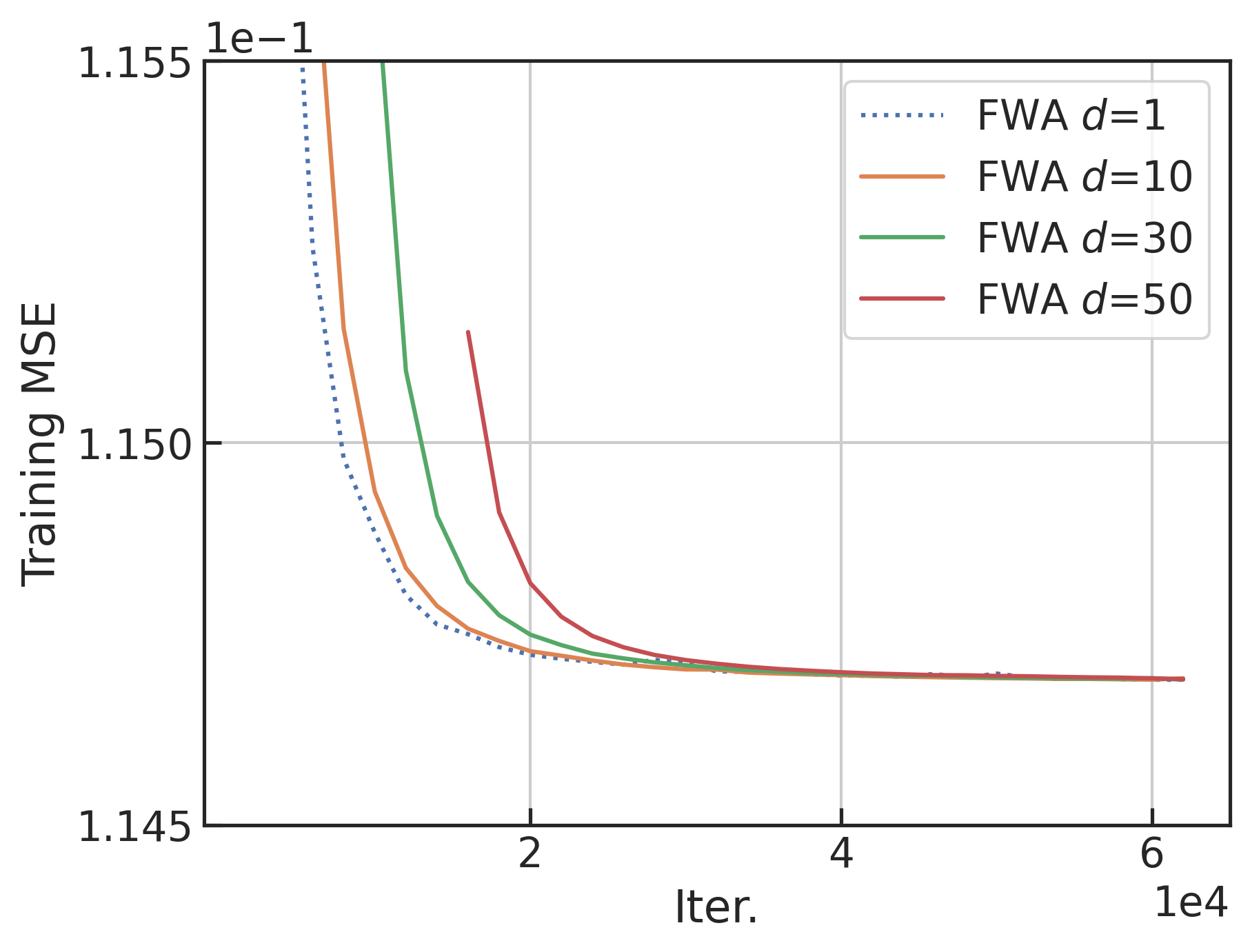}
    \caption{$K = 300$}\label{fig:convex-conergence-k300}
    \end{subfigure}
    \hfill
    \begin{subfigure}{0.45\linewidth}
    \includegraphics[width=\linewidth]{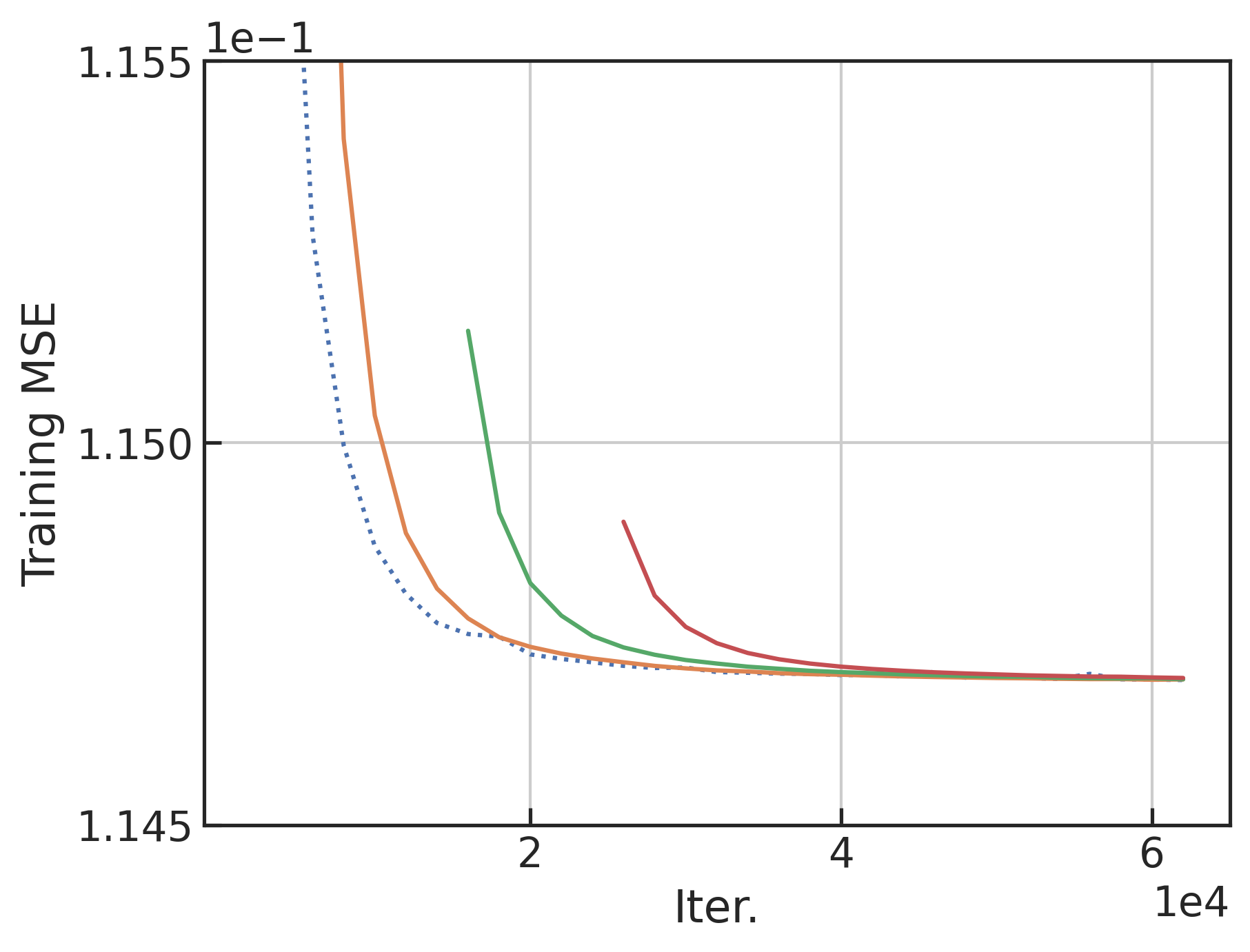}
    \caption{$K = 500$}\label{fig:convex-conergence-k500}
    \end{subfigure}
    \caption{Convergence results of the convex problem with different interval length $d$.}
    \label{fig:convex-convergence-multi-steps}
\end{figure*}

\paragraph{Results on stability and generalization.}

The results for convex and non-convex problems are shown in \Cref{fig:con-gen,fig:con-gen-cifar} respectively.
As we can see, FWA generally yields smaller parameter differences, with the effect becoming more pronounced the larger $k$ is. SGD shows the worst effect when $k=1$ and FWA performs best when $k=T$, which satisfies our theoretical results in \Cref{thm:stability-con-with,thm:stability-non-with} for convex and non-convex settings respectively. These observations are similar in the generalization error.
Due to the relative simplicity of the learning task, there is only a weak improvement in the figures when $k$ is small compared to SGD, and the effect is more obvious when $k$ is assumed to be large. In summary, the experiments coincide with our theoretical analysis, i.e., FWA achieves better generalization than vanilla SGD optimizer.
Moreover, being consistent with our analysis in \Cref{thm:stability-non-with-decay}, the cases with learning rate decay in \Cref{fig:con-gen-cifar-2} can lead to larger generalization errors, which are less stable.


\paragraph{Results on convergence.}

Finally, we evaluate the convergence analysis of convex problems in \Cref{sec:convergence}.
The convergence results are shown \Cref{fig:convex-convergence}.
For this experiment, we set the learning rate as $0.3$. For the decaying case, the learning rate is decayed from $0.6$ to $0.3$ with four steps, which is the same with previous settings. And we report the FWA with $100, 500$ and $1000$ averaging.
As we can see, 
FWA has a faster convergence rate compared to SGD, and the larger $k$ is, the faster the convergence rate is, which validates our theoretical results in \Cref{thm:convergence-cor,thm:convergence-sgd}. When $k=T$, as the purple line shows, it converges significantly slower than the others, which implies averaging hinders the convergence rate. This phenomenon is intuitive that there is a high probability that some of the initial points are far from the minima, and including them as part of the average will not help convergence, which leads us to focus on $k$ being finite. A similar discussion appears in \cite{kaddour2022stop}. 

Additionally, we plot the convergence results when the average interval $d > 1$ in \Cref{fig:convex-convergence-multi-steps}. When the interval $d$ increases, the convergence becomes slower, which is consistent with results in \Cref{thm:convergence-LAWA}.



\section{Conclusions}
We first generalize SGD and LAWA to the finite averaging problem (FWA) and obtain its convergence rate based on the proof skill of convergence for the last iteration. Moreover, we have established generalization bounds for FWA based on stability in different cases. Based on this, by comparing the SGD algorithm and further analysis, we theoretically show that FWA can achieve faster convergence and better generalization compared with SGD. For both optimization and generalization, our results can degenerate to those of SGD when $k=1$, demonstrating our method's generality. In generalization, we explain why FWA performs well: The cumulative gradient significantly affects averaging algorithms. Adopting a smaller learning rate during the final averaging stage can result in a reduced cumulative gradient and better generalization performance. Then, in the non-convex case, the expansion caused by FWA reduces to the original $\frac{1}{k}$. Additionally, we explained why early stopping can improve generalization by providing a stability-based bound.

\textbf{Limitation.} 
The theoretical analysis of algorithms is carried out under some classical assumptions, such as $L$-Lipschitz and $\beta$-smooth, which seem to be standard assumptions in the analysis of stability and generalization. However, they may be restricted in practice. Many recent studies have tried removing these assumptions in algorithmic convergence \cite{nguyen2021unified} and stability analysis \cite{lei2020fine}. It would be an exciting direction to enhance our work in the future further.
Current results mainly focus on uniformly weighted average methods, which exclude the EMA approach. In the future, we may extend our work to the weighted average setting to cover more averaging-type methods and characterize the convergence and generalization properties. 




\acks{This work is supported by STI 2030—Major Projects (No. 2021ZD0201405), the National Natural Science Foundation of China under grant 12171178. Dr. Tao's research is partially supported by NTU RSR and Start-Up Grants.}


\newpage

\appendix

\section{Proof of Some Basic Properties}
\subsection{Proof of FWA's update rules.}\label{pro-FWA-update}
According to the definition of FWA in Eq.~\eqref{FWA-rules}, we have 
\begin{equation}
\begin{aligned}
    \bar{w}^{k}_{T} &= \frac{1}{k}\left(\rho_1 w_{T-k+1} + \rho_2 w_{T-k+2} + \cdots + \rho_k w_{T}\right) \\
    \bar{w}^{k}_{T-1} &= \frac{1}{k}\left(\rho_1 w_{T-k} + \rho_2 w_{T-k+1} + \cdots + \rho_k w_{T-1}\right).    
\end{aligned}
\end{equation}
Then, we can get recursively 
\begin{equation}
\begin{aligned}
    \bar{w}^{k}_{T} - \bar{w}^{k}_{T-1} &= \frac{1}{k}\left(\rho_1 (w_{T-k+1} - w_{T-k}) + \rho_2 (w_{T-k+2} - w_{T-k+1}) + \cdots + \rho_k (w_{T} - w_{T-1})\right) \\
    &= \frac{1}{k}(\rho_1 \alpha_{T-k+1} \nabla F(w_{T-k},z_{T-k+1}) + \rho_2 \alpha_{T-k+2} \nabla F(w_{T-k+1},z_{T-k+2}) + \\
    &\cdots + \rho_k \alpha_T \nabla F(w_{T-1},z_T)),   
\end{aligned}
\end{equation}
where the second equality follows from the update of SGD. Therefore, it is not difficult to find the relationship between $\bar{w}^{k}_{T}$ and $\bar{w}^{k}_{T-1}$, i.e.,
\begin{equation}
    \bar{w}^{k}_{T} = \bar{w}^{k}_{T-1} - \frac{1}{k}\sum_{i=T-k+1}^{T} \rho_{i-(T-k)}\alpha_i\nabla F(w_{i-1},z_i).
\end{equation}

\subsection{Proof of Lemma \ref{lemma}}\label{pro-lemma}  
\paragraph{\emph{Non}-expansive.} Function is convexity and $\beta$-smoothness that implies 
\begin{equation}\label{eq:app2.2}
     \begin{aligned}
      \langle \nabla F(w) -\nabla F(v), w - v \rangle \geq \frac{1}{\beta} \Vert \nabla F(w) -\nabla F(v)\Vert^2 .
     \end{aligned}
\end{equation}
We conclude that
\begin{equation}\label{eq:app2.3}
     \begin{aligned}
      &\Vert w_{T+1} - w_{T+1}^{\prime}\Vert = \sqrt{\Vert w_{T} - \alpha \nabla F(w_{T}) - w_{T}^{\prime} + \alpha \nabla F(w_{T}^{\prime})\Vert^2}  \\
      &=\sqrt{\Vert w_{T} - w_{T}^{\prime} \Vert^2 - 2\alpha\langle \nabla F(w_{T}) -\nabla F(w_{T}^{\prime}), w_T- w_T^{\prime} \rangle +\alpha^2 \Vert \nabla F(w_{T}) - \nabla F(w_{T}^{\prime})\Vert^2} \\
      &\leq \sqrt{\Vert w_T- w_T^{\prime}\Vert^2 - \left(\frac{2\alpha}{\beta} -\alpha^2 \right) \Vert \nabla F(w_{T}) -\nabla F(w_{T}^{\prime})\Vert^2} \\
      &\leq \Vert w_T- w_T^{\prime}\Vert.
     \end{aligned}
\end{equation}

\paragraph{\boldmath$(1+\alpha\beta)$-expansive.} According to triangle inequality and $\beta$-smoothness,
\begin{equation}\label{eq:app2.1}
     \begin{aligned}
      \Vert w_{T+1} - w_{T+1}^{\prime}\Vert &\leq \Vert w_T- w_T^{\prime}\Vert + \alpha\Vert \nabla g(w_T) -\nabla g(w_T^{\prime})\Vert \\
      &\leq \Vert w_T- w_T^{\prime}\Vert + \alpha\beta \Vert w_T- w_T^{\prime}\Vert \\
      &= (1+\alpha\beta)\Vert w_T- w_T^{\prime}\Vert .
     \end{aligned}
\end{equation}

\section{Proof of the generalization bounds}\label{pro-con}
As we all know, there is $\Vert \bar{w}^{k}_{T} - \bar{w}^{k\prime}_{T}\Vert \leq \frac{1}{k}\sum_{i=T-k+1}^{T}\Vert w_{i}-w_{i}^{\prime}\Vert$ based on the triangle inequality. For convenience, we will use the notation $\bar{\delta}_{T}$, where $\bar{\delta}_{T} = \frac{1}{k}\sum_{i=T-k+1}^{T}\Vert w_{i}-w_{i}^{\prime}\Vert$, in the proof of the generalization section.

\subsection{\textbf{Proof of Theorem \ref{thm:stability-con-with}}}\label{proof-thm-con-with}
  First, using the relationship between $\bar{w}^{k}_{T}$ and $\bar{w}^{k}_{T-1}$ in Eq.~\eqref{lemma-FWA-update}, we consider that the different sample $z_{T}$ and $z_{T}^{\prime}$ are selected to update with probability $\frac{1}{n}$ at the step $T$.  
\begin{equation}
  \begin{aligned}
   \bar{\delta}_{T} &\leq \bar{\delta}_{T-1} + \frac{1}{k}\sum_{i=T-k+1}^{T} \rho_{i-(T-k)}\alpha_i \Vert\nabla F(w^{\prime}_{i-1},z_i) - \nabla F(w_{i-1},z_i) \Vert \\
   &\leq \bar{\delta}_{T-1} + \frac{2\rho_k\alpha_T L}{k} + \frac{1}{k}\sum_{i=T-k+1}^{T-1} \rho_{i-(T-k)}\alpha_i \Vert\nabla F(w^{\prime}_{i-1},z_i) - \nabla F(w_{i-1},z_i) \Vert ,
  \end{aligned}
 \end{equation}
where the proof follows from the Eq.~\eqref{lemma-FWA-update}, triangle inequality and $L$-Lipschitz condition. For $\frac{1}{k}\sum_{i=T-k+1}^{T-1} \rho_{i-(T-k)}\alpha_i \Vert\nabla F(w^{\prime}_{i-1},z_i) - \nabla F(w_{i-1},z_i) \Vert$ will be controlled in the late.

Second, another situation needs to be considered in case of the same sample is selected $(z_{T}=z_{T}^{\prime})$ to update with probability $1-\frac{1}{n}$ at the step $T$. 
\begin{equation}
  \begin{aligned}
   \bar{\delta}_{T} &\leq \bar{\delta}_{T-1} + \frac{1}{k}\sum_{i=T-k+1}^{T-1} \rho_{i-(T-k)}\alpha_i\Vert\nabla F(w^{\prime}_{i-1},z_i) - \nabla F(w_{i-1},z_i) \Vert ,
  \end{aligned}
 \end{equation}
where the second inequality follows from the non-expansive property of the convex function.

For each $\Vert\nabla F(w^{\prime}_{i-1},z_i)-\nabla F(w_{i-1},z_i)\Vert$ in the sense of expectation, We consider two situations using $\alpha L$ bound and the non-expansive property. Then, we get  
  \begin{equation}
    \frac{1}{k}\sum_{i=T-k+1}^{T-1}\rho_{i-(T-k)}\alpha_i \Vert\nabla F(w^{\prime}_{i-1},z_i) - \nabla F(w_{i-1},z_i) \Vert \leq \frac{2L}{nk}\sum_{i=T-k+1}^{T-1}\rho_{i-(T-k)}\alpha_i.
 \end{equation}

Then, we obtain the expectation based on the above analysis 
  \begin{equation}
  \begin{aligned}
    \mathbb{E}\left[\bar{\delta}_{T}\right] &\leq (1-\frac{1}{n})\bar{\delta}_{T-1} + \frac{1}{n}\left(\bar{\delta}_{T-1}+\frac{2\rho_{k}\alpha_T L}{k}\right) + \frac{2L}{nk}\sum_{i=T-k+1}^{T-1}\rho_{i-(T-k)} \alpha_i\\
    &\leq \mathbb{E}\left[\bar{\delta}_{T-1}\right] + \frac{2L}{nk}\sum_{i=T-k+1}^{T}\rho_{i-(T-k)}\alpha_i
  \end{aligned}
 \end{equation}
recursively, we can get 
    \begin{equation}
     \begin{aligned}
      \mathbb{E}\left[\bar{\delta}_{T}\right]&\leq \frac{2L}{nk} \left( \sum_{i=T-k+1}^{T}\rho_{i-(T-k)}\alpha_i + \sum_{i=T-k}^{T-1}\rho_{i-(T-1-k)}\alpha_i + \cdots  + \sum_{i=2}^{k+1}\rho_{i-1}\alpha_i \right) \\ & + \frac{2L}{nk} \left( \sum_{i=1}^{k}\rho_{i}\alpha_i + \sum_{i=1}^{k-1}\rho_{i}\alpha_i + \cdots + \sum_{i=1}^{1}\rho_{i}\alpha_i \right). \\
     \end{aligned}
    \end{equation}
Plugging this back into Eq.~\eqref{convex-basic}, we obtain
 \begin{equation}
  \epsilon_{gen} = \mathbb{E}\vert F(\bar{w}_E;z)-F(\bar{w}^{\prime}_E;z)\vert \leq \frac{2L^2}{nk} \left( \sum_{t=1}^{k}\sum_{i=1}^{t} \rho_i \alpha_i + \sum_{t=k+1}^{T}\sum_{i=t-k+1}^{t} \rho_{i-(t-k)} \alpha_i \right).
 \end{equation}
And we finish the proof.

\subsection{Proof of Eq.~\ref{nonconvex-basic}}\label{proof-noncon-basic}
We consider that $S$ and $S^\prime$ are two samples of size $n$ differing in only a single example. Let $\xi$ denote the event $\bar{\delta}_{t_0}=0$. Let $z$ be an arbitrary example and consider the random variable $I$ assuming the index of the first time step using the different sample. then we have
    \begin{equation}
     \begin{aligned}
      \mathbb{E}\vert \nabla F(\bar{w}_T^{k};z)-\nabla F(\bar{w}^{k\prime}_T;z)\vert &= P\left\lbrace \xi\right\rbrace \mathbb{E}[\vert \nabla F(\bar{w}_T^{k};z)-\nabla F(\bar{w}^{k\prime}_T;z)\vert|\xi]\\
      &+P\left\lbrace \xi^{c}\right\rbrace E[\vert \nabla F(\bar{w}_T^{k};z)-\nabla F(\bar{w}^{k\prime}_T;z)\vert |\xi^{c}]\\
      &\leq P\left\lbrace I\geq t_0\right\rbrace \cdot \mathbb{E}[\vert \nabla F(\bar{w}_T^{k};z)-\nabla F(\bar{w}^{k\prime}_T;z)\vert |\xi] \\
      &+P\left\lbrace I\leq t_0\right\rbrace \cdot \mathop{sup}_{\bar{w}^{k},z} F(\bar{w}^{k};z),\\
     \end{aligned}
    \end{equation}
where $\xi^{c}$ denotes the complement of $\xi$.   

Note that Note that when $I\geq t_0$, then we must have that $\bar{\delta}_{t_0}=0$, since the execution on $S$ and $S^{\prime}$ is identical until step $t_0$. We can get $LE[\Vert\bar{w}_{T}^{k} - \bar{w}_{T}^{k\prime}\Vert|\xi]$ combined the Lipschitz continuity of $F$. Furthermore, we know $P\left\lbrace \xi^{c}\right\rbrace=P\left\lbrace \bar{\delta}_{t_0}=0\right\rbrace\leq P\left\lbrace I\leq t_0\right\rbrace$, for the random selection rule, we have 
    \begin{equation}
     \begin{aligned}
      P\left\lbrace I\leq t_0\right\rbrace \leq \sum_{t=1}^{t_0} P\left\lbrace I = t_0\right\rbrace = \frac{t_0}{n}.
     \end{aligned}
    \end{equation}
We can combine the above two parts and $F \in [0,1]$ to derive 
the stated bound $L\mathbb{E}[\Vert\bar{w}_{T}^{k} - \bar{w}_{T}^{k\prime}\Vert\vert\xi]+\frac{t_0}{n}$, which completes the proof.

\subsection{Proof of Lemma \ref{Lemma_noncon}}\label{proof-Lemma_noncon}
By triangle inequality and our assumption that $F$ satisfies, we have
 \begin{equation}
  \begin{aligned}
   \Vert w^{\prime}_{T}& - w_{T} \Vert = \frac{1}{k} \cdot k \cdot \Vert w^{\prime}_{T} - w_{T} \Vert \\
   \leq & \frac{1}{k} ( \Vert w^{\prime}_{T} - w_{T} \Vert + (1+\alpha_{T-1}\beta)\Vert w^{\prime}_{T-1} - w_{T-1} \Vert + \cdots + \\&(1+\alpha_{T-1}\beta)(1+\alpha_{T-2}\beta)\cdots(1+\alpha_{T-k+1}\beta)\Vert w^{\prime}_{T-k+1} - w_{T-k+1} \Vert ) \\
   \leq & \prod_{t=T-k+1}^{T} (1+\alpha_t\beta)\left(\frac{1}{k} \sum_{i=T-k+1}^{T}\Vert w^{\prime}_{i} - w_{i} \Vert\right).
    \end{aligned}
 \end{equation}
Let $\alpha_t = \frac{c}{t}$, we have
  \begin{equation}
  \begin{aligned}
   \Vert w^{\prime}_{T}& - w_{T} \Vert \leq &\prod_{t=T-k+1}^{T} (1+\alpha_t\beta)\bar{\delta}_{T} \leq \left(1+ \frac{c\beta}{T-k}\right)^k\bar{\delta}_{T} \leq e^\frac{c\beta k}{T-k}\bar{\delta}_{T}.
  \end{aligned}
 \end{equation}
Let $\alpha_t = \alpha = \frac{c}{T}$, we have
  \begin{equation}
  \begin{aligned}
   \Vert w^{\prime}_{T}& - w_{T} \Vert \leq &\prod_{t=T-k+1}^{T} (1+\alpha_t\beta)\bar{\delta}_{T} \leq \left(1+ \frac{c\beta}{T}\right)^k\bar{\delta}_{T} \leq e^\frac{c\beta k}{T}\bar{\delta}_{T},
  \end{aligned}
 \end{equation}
where $\bar{\delta}_{T}=\frac{1}{k}\sum_{i=T-k+1}^{T}\Vert w_{i} - w_{i}^{\prime} \Vert$.

\subsection{\textbf{Proof. Theorem \ref{thm:stability-non-with} (Based on the constant learning rate)}}
\label{proof-thm-non-with} In the case of non-convex, the $(1+\alpha\beta)$-expansive properties and $L$-Lipschitz conditions are used in our proof. Based on the relationship between $\bar{w}^{k}_T$ and $\bar{w}^{k}_{T-1}$ \ref{lemma-FWA-update}. We consider that the different samples $z_T$ and $z^{\prime}_T$ are selected to update with probability $\frac{1}{n}$ at step T.
\begin{equation}
  \begin{aligned}
   \bar{\delta}_{T} &\leq \bar{\delta}_{T-1} + \frac{1}{k}\sum_{i=T-k+1}^{T} \alpha \Vert\nabla F(w^{\prime}_{i-1},z_i) - \nabla F(w_{i-1},z_i) \Vert \\
   &\leq \bar{\delta}_{T-1} + \frac{2\alpha L}{k} + \frac{1}{k}\sum_{i=T-k+1}^{T-1} \alpha \Vert\nabla F(w^{\prime}_{i-1},z_i) - \nabla F(w_{i-1},z_i) \Vert ,
  \end{aligned}
 \end{equation}
Next, the same sample $z=z^{\prime}$ is selected to update with probability $1-\frac{1}{n}$ at step T.
\begin{equation}
  \begin{aligned}
   \bar{\delta}_{T} &\leq \bar{\delta}_{T-1} + \frac{1}{k}\sum_{i=T-k+1}^{T} \alpha \Vert\nabla F(w^{\prime}_{i-1},z_i) - \nabla F(w_{i-1},z_i) \Vert \\
   &\leq \bar{\delta}_{T-1} + \frac{\alpha \beta}{k}\Vert w^{\prime}_{T-1} - w_{T-1} \Vert + \frac{1}{k}\sum_{i=T-k+1}^{T-1} \alpha \Vert\nabla F(w^{\prime}_{i-1},z_i) - \nabla F(w_{i-1},z_i) \Vert \\
   &\leq (1+\frac{\alpha \beta(1+\alpha \beta)^{k-1}}{k})\bar{\delta}_{T-1} + \frac{1}{k}\sum_{i=T-k+1}^{T-1} \alpha \Vert\nabla F(w^{\prime}_{i-1},z_i) - \nabla F(w_{i-1},z_i) \Vert,
  \end{aligned}
 \end{equation}
where the proof follows from the $\beta$-smooth and Lemma \ref{Lemma_noncon}. 

Then, we bound the $\alpha \Vert\nabla F(w^{\prime}_{T-2},z_{T-1}) - \nabla F(w_{T-2},z_{T-1}) \Vert$ with different sampling. 
 \begin{equation}\label{noncon-sigbound}
  \begin{aligned}    
    \alpha\Vert\nabla &F(w^{\prime}_{T-2},z_{T-1}) - \nabla F(w_{T-2},z_{T-1}) \Vert = \frac{2\alpha L}{n} + \left(1-\frac{1}{n}\right)\alpha\beta\Vert w_{T-2} - w^{\prime}_{T-2} \Vert\\
    &\leq \frac{2\alpha L}{n} + \alpha\beta\left(\Vert w_{T-3} - w^{\prime}_{T-3} \Vert + \alpha \Vert \nabla F(w^{\prime}_{T-3},z_{T-2})-\nabla F^{\prime}(w_{T-3},z_{T-2}) \Vert\right) \\
    &\leq \frac{2\alpha L}{n} + \alpha\beta\left(\frac{2\alpha L}{n} + (1+\alpha\beta)\Vert w_{T-3} - w^{\prime}_{T-3} \Vert\right) \\
    &\cdots\\
    &\leq \frac{2\alpha L}{n}(1+\alpha\beta)^{T-2-t_0} + \alpha\beta(1+\alpha\beta)^{T-2-t_0}\Vert w_{t_{0}} - w^{\prime}_{t_{0}} \Vert =\frac{2\alpha L}{n}(1+\alpha\beta)^{T-2-t_0},
  \end{aligned}
 \end{equation}
where $w_{t_{0}} = w^{\prime}_{t_{0}}$. Therefore, we can obtain the bound for $\frac{1}{k}\sum_{i=T-k+1}^{T-1} \alpha \Vert\nabla F(w^{\prime}_{i-1},z_i) - \nabla F(w_{i-1},z_i) \Vert$ in the expectation sense.
\begin{equation}\label{44}
    \begin{aligned}
     \frac{\alpha}{k}\sum_{i=T-k+1}^{T-1} & \mathbb{E} \Vert\nabla F(w^{\prime}_{i-1},z_i) - \nabla F(w_{i-1},z_i) \Vert 
     \leq \frac{2\alpha L}{nk} \sum_{i=T-k}^{T-2} (1+\alpha\beta)^{i-t_{0}} \\
     &\leq \frac{2\alpha L}{nk} \cdot k (1+\alpha\beta)^{T} \leq \frac{2\alpha L(1+\alpha\beta)^{T}}{n}.  
    \end{aligned}
\end{equation}
Then, we obtain the expectation considering the above analysis 
 \begin{equation}
    \begin{aligned}
     \mathbb{E}\left[\bar{\delta}_{T+1}\right] &\leq (1-\frac{1}{n})\left(1+\frac{\alpha \beta(1+\alpha \beta)^{k-1}}{k}\right)\bar{\delta}_T + \frac{1}{n}\left(\bar{\delta}_T+\frac{2\alpha L}{k}\right) + \frac{2\alpha L(1+\alpha\beta)^{T}}{n}\\ 
     &\leq \left(\frac{1}{n}+(1-\frac{1}{n})\left(1+\frac{\alpha \beta(1+\alpha \beta)^{k-1}}{k}\right)\right)\bar{\delta}_{T} + \frac{2\alpha L}{nk}\left(1+k(1+\alpha\beta)^{T}\right)\\
    \end{aligned}
   \end{equation}
let $\alpha = \frac{c}{t}$, then
  \begin{equation}
 \begin{aligned}
     &= \left(1+(1-\frac{1}{n})\frac{c\beta(1+\frac{c\beta}{t})^{k}}{kt}\right) \bar{\delta}_{t} + \frac{2cL}{nkt}\left(1+k(1+\frac{c\beta}{t})^{t}\right)\\
     &\leq \exp\left((1-\frac{1}{n})\frac{c\beta e^{\frac{c\beta k}{t}}}{kt}\right) \bar{\delta}_{t} + \frac{2cL}{kn}\cdot\frac{1+k e^{c\beta}}{t}.
    \end{aligned}
   \end{equation}
Here we used that $\lim\limits_{x\to\infty}(1+\frac{1}{x})^x=e$ and $\lim\limits_{x\to\infty}e^\frac{1}{x}=1$. 
Using the fact that $\bar{\delta}_{t_0}=0$, we can unwind this recurrence relation from $T$ down to $t_0+1$.
  \begin{equation}
    \begin{aligned}
     \mathbb{E}\bar{\delta}_{t} &\leq \sum_{t=t_0 +1}^{T} \left( \prod_{m=t+1}^{T}\exp\left((1-\frac{1}{n})\frac{c\beta}{km}\right)\right)\frac{2cL}{kn}\cdot\frac{1+k e^{c\beta}}{t}\\
     &= \sum_{t=t_0 +1}^{T} \exp\left(\frac{(1-\frac{1}{n})c\beta}{k} \sum_{m=t+1}^{T}\frac{1}{m}\right)\frac{2cL}{kn}\cdot\frac{1+k e^{c\beta}}{t}\\
     &\leq \sum_{t=t_0 +1}^{T} \exp\left( \frac{(1-\frac{1}{n})c\beta}{k} \cdot \log(\frac{T}{t}) \right)\frac{2cL}{kn}\cdot\frac{1+k e^{c\beta}}{t}\\
     &\leq T^{\frac{(1-\frac{1}{n})c\beta}{k}} \cdot \sum_{t=t_0 +1}^{T} \left(\frac{1}{t}\right)^{\frac{(1-\frac{1}{n})c\beta}{k}+1} \cdot \frac{2cL(1+ke^{c\beta})}{kn}\\
     &\leq \frac{k}{(1-\frac{1}{n})c\beta} \cdot \frac{2cL(1+ke^{c\beta})}{kn} \cdot \left(\frac{T}{t_0}\right)^{\frac{(1-\frac{1}{n})c\beta}{k}}\\
     &\leq \frac{2L(1+ke^{c\beta})}{(n-1)\beta} \cdot \left(\frac{T}{t_0}\right)^{\frac{c\beta}{k}}.
    \end{aligned}
   \end{equation}
Plugging this back into Eq.~\eqref{nonconvex-basic}, we obtain
 \begin{equation}\label{with-con}
  \mathbb{E}|F(\bar{w}_T;z)-F(\bar{w}^{\prime}_T;z)| \leq \frac{t_0}{n} + \frac{2L^2(1+ke^{c\beta})}{(n-1)\beta} \cdot \left(\frac{T}{t_0}\right)^{\frac{c\beta}{k}}.
 \end{equation}
By taking the extremum, we obtain the minimum  
 \begin{equation}\label{with-con-t_0}
    t_0 = \left(2cL^2(1+ke^{c\beta})k^{-1}\right)^{\frac{k}{c\beta+k}}\cdot T^{\frac{c\beta}{c\beta+k}}
   \end{equation}
finally, this setting gets
 \begin{equation}\label{with-con-result}
  \epsilon_{gen} = \mathbb{E}\vert F(\bar{w}_T;z)-F(\bar{w}^{\prime}_T;z)\vert \leq \frac{1+\frac{1}{c\beta}}{n-1}\left(2cL^2(1+ke^{c\beta})k^{-1}\right)^{\frac{k}{c\beta+k}}\cdot T^{\frac{c\beta}{c\beta+k}},
 \end{equation}
to simplify, omitting constant factors that depend on $\beta$, c
and L, we get 
   \begin{equation}
    \epsilon_{gen}  \leq \mathcal{O}\left(\frac{T^{\frac{c\beta}{c\beta+k}}}{n}\right).
   \end{equation}
And we finish the proof.

\subsection{\textbf{Proof. Theorem \ref{thm:stability-non-with-decay} (Based on decaying learning rate)}}\label{proof-thm-non-with-decay} 
In the case of non-convex, the $(1+\alpha\beta)$-expansive properties and $L$-Lipschitz conditions are used in our proof. Based on the relationship between $\bar{w}^{k}_T$ and $\bar{w}^{k}_{T-1}$ \ref{lemma-FWA-update}. We consider that the different samples $z_T$ and $z^{\prime}_T$ are selected to update with probability $\frac{1}{n}$ at step T.
\begin{equation}
  \begin{aligned}
   \bar{\delta}_{T} &\leq \bar{\delta}_{T-1} + \frac{1}{k}\sum_{i=T-k+1}^{T} \rho_{i-(T-k)} \alpha_{i} \Vert\nabla F(w^{\prime}_{i-1},z_i) - \nabla F(w_{i-1},z_i) \Vert \\
   &\leq \bar{\delta}_{T-1} + \frac{2 \rho_{k} \alpha_{T} L}{k} + \frac{1}{k}\sum_{i=T-k+1}^{T-1} \rho_{i-(T-k)} \alpha_{i} \Vert\nabla F(w^{\prime}_{i-1},z_i) - \nabla F(w_{i-1},z_i) \Vert ,
  \end{aligned}
 \end{equation}
 
Next, the same sample $z=z^{\prime}$ is selected to update with probability $1-\frac{1}{n}$ at step T.
\begin{equation}
  \begin{aligned}
   \bar{\delta}_{T} &\leq \bar{\delta}_{T-1} + \frac{1}{k}\sum_{i=T-k+1}^{T} \rho_{i-(T-k)} \alpha_{i} \Vert\nabla F(w^{\prime}_{i-1},z_i) - \nabla F(w_{i-1},z_i) \Vert \\
   &\leq \bar{\delta}_{T-1} + \frac{\rho_k \alpha_{T} \beta}{k}\Vert w^{\prime}_{T-1} - w_{T-1} \Vert + \frac{1}{k}\sum_{i=T-k+1}^{T-1} \rho_{i-(T-k)} \alpha_{i} \Vert\nabla F(w^{\prime}_{i-1},z_i) - \nabla F(w_{i-1},z_i) \Vert \\
   &\leq (1+\frac{\rho_{k} \alpha_{T} \beta e^{\frac{c\beta k}{T-k}}}{k})\bar{\delta}_{T-1} + \frac{1}{k}\sum_{i=T-k+1}^{T-1} \rho_{i-(T-k)} \alpha_{i} \Vert\nabla F(w^{\prime}_{i-1},z_i) - \nabla F(w_{i-1},z_i) \Vert,
  \end{aligned}
 \end{equation}
where the proof follows from the $\beta$-smooth and Lemma \ref{Lemma_noncon}. 

Then, we bound the $\frac{1}{k} \sum_{i=T-k+1}^{T-1} \rho_{i-(T-k)}\alpha_i \Vert\nabla F(w^{\prime}_{i-1},z_i) - \nabla F(w_{i-1},z_i) \Vert$ in the sense of expectation. For the each term $\rho_{i-(T-k)} \alpha_{i} \Vert\nabla F(w^{\prime}_{i},z_{i+1}) - \nabla F(w_{i},z_{i+1}) \Vert$, we have the recursive relationship
 \begin{equation}\label{noncon-sigbound}
  \begin{aligned}    
    \rho_{i-(T-k)}&\alpha_{i}\mathbb{E}\Vert\nabla F(w^{\prime}_{i},z_{i+1}) - \nabla F(w_{i},z_{i+1}) \Vert = \frac{2\rho_{i-(T-k)}\alpha_{i} L}{n} + \left(1-\frac{1}{n}\right)\rho_{i-(T-k)}\alpha_{i}\beta\Vert w_{i} - w^{\prime}_{i} \Vert\\
    &\leq \frac{2\rho_{i-(T-k)}\alpha_i L}{n} \! + \! \rho_{i-(T-k)}\alpha_{i}\beta\left(\Vert w_{i-1} - w^{\prime}_{i-1} \Vert \!+\! \alpha_{i-1} \Vert \nabla F(w^{\prime}_{i-1},z_{i})-\nabla F^{\prime}(w_{i-1},z_{i}) \Vert\right) \\
    &\leq \frac{2\rho_{i-(T-k)}\alpha_i L}{n} + \rho_{i-(T-k)}\alpha_i\beta\left(\frac{2\alpha_{i-1} L}{n} + (1+\alpha_{i-1}\beta)\Vert w_{i-1} - w^{\prime}_{i-1} \Vert\right) \\
    &\cdots\\
    &\leq \frac{2\rho_{i-(T-k)}\alpha_i L}{n} \left(1+\alpha_{i-1}\beta + \sum_{m=t_0}^{i-1}\prod_{t=m+1}^{i}(1+\alpha_t\beta)\alpha_m\right) \\
    & + \rho_{i-(T-k)}\alpha_i\beta\prod_{t=t_0}^{i}(1+\alpha_t\beta)\Vert w_{t_0} - w^{\prime}_{t_0} \Vert,
  \end{aligned}
 \end{equation}
where the second term $\rho_{i-(T-k)}\alpha_i\beta\prod_{t=t_0}^{i}(1+\alpha_t\beta)\Vert w_{t_0} - w^{\prime}_{t_0} \Vert =0$ according to $w_{t_0} = w^{\prime}_{t_0}$. 
Therefore, we discuss the bound for $\frac{1}{k}\sum_{i=T-k+1}^{T-1} \rho_{i-(T-k)}\alpha_{i} \Vert\nabla F(w^{\prime}_{i-1},z_i) - \nabla F(w_{i-1},z_i) \Vert$ based on the recursive relationship.
\begin{equation}\label{noncon-with-bound}
    \begin{aligned}
     & \frac{1}{k}\sum_{i=T-k+1}^{T-1} \rho_{i-(T-k)}\alpha_{i} \mathbb{E} \Vert\nabla F(w^{\prime}_{i-1},z_i) - \nabla F(w_{i-1},z_i) \Vert  \\
     & \leq \frac{1}{k}\sum_{i=T-k+1}^{T-1} \frac{2\rho_{i-(T-k)}\alpha_i L}{n}\! \left(\!\!1+\alpha_{i-1}\beta + \sum_{m=t_0}^{i-1}\prod_{t=m+1}^{i}(1+\alpha_t\beta)\alpha_m \!\right) \\
     & = \frac{2L}{k}\sum_{i=T-k+1}^{T-1} \frac{\rho_{i-(T-k)}\alpha_i}{n} + \frac{2\beta L}{k}\sum_{i=T-k+1}^{T-1} \frac{\rho_{i-(T-k)}\alpha_i \alpha_{i-1}}{n} \\
     & +\frac{2L}{k}\sum_{i=T-k+1}^{T-1} \frac{\rho_{i-(T-k)}\alpha_i}{n} \sum_{m=t_0}^{i-1}\prod_{t=m+1}^{i}(1+\alpha_t\beta)\alpha_{m} ,  
    \end{aligned}
\end{equation}
Let $\alpha_t=\frac{c}{t}$, the first term of last line in Eq.~\eqref{noncon-with-bound}  
\begin{equation}
    \begin{aligned}
     \frac{2L}{k}\sum_{i=T-k+1}^{T-1} \frac{\rho_{i-(T-k)}\alpha_i}{n} = \frac{2cL}{nk} \sum_{i=T-k+1}^{T-1} \frac{\rho_{i-(T-k)}}{i} \leq \frac{2cL}{nk(T-k+1)}\cdot \sum_{i=T-k+1}^{T-1}\rho_{i-(T-k)}.
    \end{aligned}
\end{equation}
The second term of the last line in Eq.~\eqref{noncon-with-bound}  
\begin{equation}
    \begin{aligned}
     \frac{2\beta L}{k}\sum_{i=T-k+1}^{T-1} \frac{\rho_{i-(T-k)}\alpha_i \alpha_{i-1}}{n} = \frac{2\beta c^2L}{nk} \sum_{i=T-k+1}^{T-1} \frac{\rho_{i-(T-k)}}{i(i-1)} \leq \frac{2c^2\beta L}{nk(T-k)^2}\cdot\sum_{i=T-k+1}^{T-1}\rho_{i-(T-k)}.
    \end{aligned}
\end{equation}
The last term of the last line in Eq.~\eqref{noncon-with-bound}  
\begin{equation}\label{41}
    \begin{aligned}
    \frac{2L}{k}&\sum_{i=T-k+1}^{T-1} \frac{\rho_{i-(T-k)}\alpha_i}{n} \sum_{m=t_0}^{i-1}\prod_{t=m+1}^{i}(1+\alpha_t\beta)\alpha_{m} = \frac{2cL}{nk}\sum_{i=T-k+1}^{T-1} \frac{\rho_{i-(T-k)}}{i} \sum_{m=t_0}^{i-1}\prod_{t=m+1}^{i}(1+\frac{c\beta}{t})\frac{c}{m} \\
    &\leq \frac{2c^2L\sum_{i=T-k+1}^{T-1} \rho_{i-(T-k)}}{nk(T-k+1)}\left( \sum_{m=t_0}^{T-1}\prod_{t=m}^{T-1}(1+\frac{c\beta}{t})\frac{1}{m} \right) \\
    &\leq \frac{2cL\sum_{i=T-k+1}^{T-1} \rho_{i-(T-k)}}{nk\beta}\left(\frac{1}{T-k}\right)^{1-\frac{kc\beta}{k+c\beta}}.
    \end{aligned}
\end{equation}
where $1-\frac{kc\beta}{k+c\beta}>0$. Here we provide the proof of Eq.~\eqref{41} as follows
\begin{equation}
    \begin{aligned}
    \sum_{m=t_0}^{T-1}\prod_{t=m}^{T-1}(1+\frac{c\beta}{t})\frac{1}{m} &\leq \sum_{m=t_0}^{T-1} \frac{1}{m} \left(e^{\sum_{t=m}^{T-1}\frac{c\beta}{t}}\right)\leq \sum_{m=t_0}^{T-1} \frac{T^{c\beta}}{m^{1+c\beta}}\leq T^{c\beta}\int_{t_0}^{T-1}m^{-(1+c\beta)}dm \\
    & = \frac{T^{c\beta}}{c\beta} \left(\frac{1}{t_0^{c\beta}} -\frac{1}{(T-1)^{c\beta}} \right)\leq \frac{1}{c\beta} \cdot\left(\frac{T}{t_0}\right)^{c\beta}.
    \end{aligned}
\end{equation}
Taking $t_0=T^{\frac{c\beta}{c\beta+k}}$, and we obtain the Eq.~\eqref{41}. Let $M_1=\left(1+\frac{1}{\beta}+c\beta\right)\sum_{i=T-k+1}^{T-1} \rho_{i-(T-k)}$, we can obtain the bound for $\frac{1}{k}\sum_{i=T-k+1}^{T-1} \rho_{i-(T-k)}\alpha_{i} \Vert\nabla F(w^{\prime}_{i-1},z_i) - \nabla F(w_{i-1},z_i) \Vert$ in the expectation sense.
\begin{equation}
    \begin{aligned}
     \frac{1}{k}\sum_{i=T-k+1}^{T-1} & \rho_{i-(T-k)}\alpha_{i} \mathbb{E} \Vert\nabla F(w^{\prime}_{i-1},z_i) - \nabla F(w_{i-1},z_i) \Vert 
     \leq \frac{2cLM_1}{nk}\left(\frac{1}{T-k}\right)^{1-\frac{kc\beta}{k+c\beta}}.  
    \end{aligned}
\end{equation}

Then, we obtain the expectation considering the above analysis 
 \begin{equation}
    \begin{aligned}
     \mathbb{E}\left[\bar{\delta}_{T+1}\right] &\leq (1-\frac{1}{n})\left(1+\frac{\rho_k\alpha_{T} \beta e^{\frac{c\beta k}{T-k}}}{k}\right)\bar{\delta}_T + \frac{1}{n}\left(\bar{\delta}_T+\frac{2\rho_k\alpha_T L}{k}\right) + \frac{2cLM_1}{nk}\left(\frac{1}{T-k}\right)^{1-\frac{kc\beta}{k+c\beta}}\\ 
     &\leq \left(\frac{1}{n}+(1-\frac{1}{n})\left(1+\frac{\rho_k\alpha_{T} \beta e^{\frac{c\beta k}{T-k}}}{k}\right)\right)\bar{\delta}_{T} + \frac{2\rho_k\alpha_T L}{nk}+\frac{2cLM_1}{nk}\left(\frac{1}{T-k}\right)^{1-\frac{kc\beta}{k+c\beta}}\\
    \end{aligned}
   \end{equation}
let $\alpha_t = \frac{c}{t}$, then
  \begin{equation}
 \begin{aligned}
     &= \left(1+(1-\frac{1}{n})\frac{\rho_k c\beta e^{\frac{c\beta k}{t-k}}}{kt}\right) \bar{\delta}_{t} + \frac{2cL(\rho_k+M_1)}{nk}\left(\frac{1}{t-k}\right)^{1-\frac{kc\beta}{k+c\beta}}\\
     &\leq \exp\left((1-\frac{1}{n})\frac{\rho_k c\beta }{kt}\right) \bar{\delta}_{t} + \frac{2cLM}{nk}\left(\frac{1}{t-k}\right)^{1-\frac{kc\beta}{k+c\beta}},
    \end{aligned}
   \end{equation}
where $M=\left(1+\frac{1}{\beta}+c\beta\right)\sum_{i=T-k+1}^{T} \rho_{i-(T-k)}$ and we used that $\lim\limits_{x\to\infty}(1+\frac{1}{x})^x=e$ and $\lim\limits_{x\to\infty}e^\frac{1}{x}=1$. 

Using the fact that $\bar{\delta}_{t_0}=0$ and $\rho_k \leq 1$, we can unwind this recurrence relation from $T$ down to $t_0+1$.
  \begin{equation}
    \begin{aligned}
     \mathbb{E}\bar{\delta}_{t+1} &\leq \sum_{t=t_0 +1}^{T} \left( \prod_{m=t+1}^{T}\exp\left((1-\frac{1}{n})\frac{c\beta}{km}\right)\right)\frac{2cLM}{nk}\cdot\left(\frac{1}{t-k}\right)^{1-\frac{kc\beta}{k+c\beta}}\\
     &= \sum_{t=t_0 +1}^{T} \exp\left(\frac{(1-\frac{1}{n})c\beta}{k} \sum_{m=t+1}^{T}\frac{1}{m}\right)\frac{2cLM}{nk}\cdot\left(\frac{1}{t-k}\right)^{1-\frac{kc\beta}{k+c\beta}}\\
     &\leq \sum_{t=t_0 +1}^{T} \exp\left( \frac{(1-\frac{1}{n})c\beta}{k} \cdot \log(\frac{T}{t}) \right)\frac{2cLM}{nk}\cdot\left(\frac{1}{t-k}\right)^{1-\frac{kc\beta}{k+c\beta}}\\
     &\leq T^{\frac{(1-\frac{1}{n})c\beta}{k}} \cdot \sum_{t=t_0 +1}^{T} \left(\frac{1}{t-k}\right)^{\frac{(1-\frac{1}{n})c\beta}{k}+1-\frac{kc\beta}{k+c\beta}} \cdot \frac{2cLM}{kn}\\
     &\leq \left(\frac{c\beta}{k}-\frac{kc\beta}{k+c\beta}\right)^{-1} \cdot \frac{2cLM}{n} \cdot T^{\frac{c\beta}{k}} \cdot \left(\frac{T}{t_0-k}\right)^{\frac{c\beta}{k}-\frac{kc\beta}{c\beta + k}}\\
     &\leq \frac{2cLM\tau}{n} \cdot T^{\frac{c\beta}{k}} \cdot\left(\frac{1}{t_0-k}\right)^{\frac{c\beta}{k}-\frac{kc\beta}{c\beta + k}},
    \end{aligned}
   \end{equation}
where $\tau = \left(\frac{c\beta}{k}-\frac{kc\beta}{k+c\beta}\right)^{-1}$ and $\frac{c\beta}{k}-\frac{kc\beta}{c\beta + k} > 0$, $k\in(1,\frac{1+\sqrt{1+4c\beta}}{2})$ and $c\beta \in (0,1)$.
Plugging this back into Eq.~\eqref{nonconvex-basic}, we obtain
 \begin{equation}\label{with-dec}
  \mathbb{E}|g(\bar{w}_T;z)-g(\bar{w}^{\prime}_T;z)| \leq \frac{t_0}{n} + \frac{2cL^2 M\tau}{n} \cdot T^{\frac{c\beta}{k}} \cdot\left(\frac{1}{t_0-k}\right)^{\frac{c\beta}{k}-\frac{kc\beta}{c\beta + k}}.
 \end{equation}
By taking the extremum, we obtain the minimum  
 \begin{equation}\label{with-dec-t_0}
    t_0 = \mathcal{O}\left( T^{\frac{kc\beta+c^2 \beta^2}{2kc\beta+c^2 \beta^2 +k^2 (1-c\beta)}}\right)
   \end{equation}
finally, omitting constant factors that depend on $\beta$, $c$ and $L$, this setting get
  \begin{equation}\label{with-dec-result}
    \epsilon_{stab} \leq \mathcal{O}\left(\frac{T^{\frac{kc\beta+c^2 \beta^2}{2kc\beta+c^2 \beta^2 +k^2 (1-c\beta)}}}{n} \right).
 \end{equation}
Compared the upper bound of SGD $\mathcal{O}\left(\frac{T^{\frac{c\beta}{1+c\beta}}}{n} \right)$, when $\frac{kc\beta+c^2 \beta^2}{2kc\beta+c^2 \beta^2 +k^2 (1-c\beta)}<\frac{c\beta}{1+c\beta}$, i.e., $k^2-k>\frac{c\beta}{1-c\beta}$ is satisfied, the upper bound of FWA is smaller than SGD. under the condition $c\beta \in (0,1)$ and $k \in (1,\frac{1+\sqrt{1+4c\beta}}{2})$, numerical verification shows that no $c\beta$ and $k$ satisfy condition $k^2-k>\frac{c\beta}{1-c\beta}$. This indicates that the generalization upper bound of FWA is greater than that of SGD.

\section{\textbf{Proof of the convergence bounds}}
\subsection{\textbf{Proof of the Proposition \ref{thm:convergence1}}}\label{proof-thm-convergence} 
1. On the one hand, by Jensen's inequality, we get
\begin{equation}\label{proof-3.1-left}
  \mathbb{E}[ F(\bar{w}_{T}^k) - F(w^{\star})] \leq  \frac{1}{k}\sum_{t=T-k+1}^{T} \mathbb{E}[ \rho_{t-(T-k)}F({w}_{t}) - F(w^{\star})].
 \end{equation}
2. On the other hand, by convexity of $W$, for any $w\in W$, we have
\begin{equation*}
     \begin{aligned}
  \mathbb{E}\Vert w_{T+1} - w \Vert^2 =& \mathbb{E} \Vert w_{T} - \alpha_T \nabla F(w_T)- w \Vert^2 \\
  = &\mathbb{E} \Vert w_{T} - w \Vert^2 -2 \alpha_T \mathbb{E}\langle \nabla F(w_T), w_T - w \rangle + \alpha_T^2 G^2.
    \end{aligned}
 \end{equation*}
Let $k\in {1,2,\cdots,\lfloor\frac{T}{2}\rfloor}$. Summing over all $t=T-k+1,\cdots,T$ and rearranging, we get 
 \begin{equation}\label{proof-3.1-key}
     \begin{aligned}
  \sum_{t=T-k+1}^{T} & \rho_{t-(T-k)}\mathbb{E}\langle \nabla F(w_t), w_t - w \rangle \! \leq \!\! \sum_{t=T-k+1}^{T} \frac{\rho_{t-(T-k)}}{2\alpha_t}\left( \mathbb{E}\Vert w_{t} - w \Vert^2 - \mathbb{E}\Vert w_{t+1} - w \Vert^2\right) \\
  & + \frac{G^2}{2} \!\! \sum_{t=T-k+1}^{T} \!\! \rho_{t-(T-k)}\alpha_t \\
  \leq &\frac{\rho_1 \mathbb{E}\Vert w_{T-k+1} - w \Vert^2}{2\alpha_{T-k+1}}+ \sum_{t=T-k+2}^{T} \mathbb{E}\Vert w_{t} - w \Vert^2 \left(\frac{\rho_{t-(T-k)}}{2\alpha_t}-\frac{\rho_{t-(T-k+1)}}{2\alpha_{t-1}} \right) \\
  & +\frac{G^2}{2} \sum_{t=T-k+1}^{T} \rho_{t-(T-k)}\alpha_t \\
    \end{aligned}
 \end{equation}
By convexity of $F$ based on the assumption \ref{Convex function}, we have 
\begin{equation}\label{proof-3.1}
\begin{aligned}
   \mathbb{E} [\sum_{t=T-k+1}^{T} &\rho_{t-(T-k)} (F(w_t)- F(w))] \leq \frac{\rho_1 \mathbb{E}\Vert w_{T-k+1} - w \Vert^2}{2\alpha_{T-k+1}} \\
   & + \sum_{t=T-k+2}^{T} \mathbb{E}\Vert w_{t} - w \Vert^2 \left(\frac{\rho_{t-(T-k)}}{2\alpha_t}-\frac{\rho_{t-(T-k+1)}}{2\alpha_{t-1}} \right) +\frac{G^2}{2} \sum_{t=T-k+1}^{T} \rho_{t-(T-k)}\alpha_t . \\
\end{aligned}
\end{equation}
Based on the inequality \eqref{proof-3.1-left}, we get
\begin{equation}\label{proof-prop}
\begin{aligned}
   \mathbb{E} [F(\bar{w}_T^k)- &F(w))] \leq \frac{\rho_1 \mathbb{E}\Vert w_{T-k+1} - w \Vert^2}{2k\alpha_{T-k+1}} + \sum_{t=T-k+2}^{T} \mathbb{E}\Vert w_{t} - w \Vert^2 \left(\frac{\rho_{t-(T-k)}}{2k\alpha_t}-\frac{\rho_{t-(T-k+1)}}{2k\alpha_{t-1}} \right) \\ & +\frac{G^2}{2k} \sum_{t=T-k+1}^{T} \rho_{t-(T-k)}\alpha_t . 
\end{aligned}
\end{equation}
Then, we finish the proof.

\subsection{\textbf{Proof of the Theorem \ref{thm:convergence-cor}}}\label{proof-thm-convergence-cor}
Let $\rho_i=1$, $\sup_{w,w^{\prime}\in {W}} \Vert w-w^{\prime} \Vert \leq D$, and pick $w=w_{T-k}$, we rewrite Eq.~\eqref{proof-3.1} as
\begin{equation*}
  \mathbb{E} [\sum_{t=T-k}^{T} (F(w_t)-F(w_{T-k})) ] 
  \leq \left(\frac{D^2}{2c}+cG^2\right)\frac{k+1}{\sqrt{T}}.
 \end{equation*}
Then, let $V_{k} = \frac{1}{k+1}\sum_{t=T-k}^{T} F({w}_{t})$, the above inequality implies that 
\begin{equation*}
  -\mathbb{E} [F(w_{T-k})] \leq -\mathbb{E} [V_k] + \left(\frac{D^2}{2c}+cG^2\right)\frac{1}{\sqrt{T}}.
 \end{equation*}
According to the definition of $V_{k}$, we get
\begin{equation}\label{proof-Vk}
\begin{aligned}
  \mathbb{E} [V_{k-1}] &= \frac{1}{k}((k+1)\mathbb{E} [V_{k}] -\mathbb{E} [F(w_{T-k})] )\\
  &\leq \mathbb{E} [V_{k}] + \left(\frac{D^2}{2c}+cG^2\right)\frac{1}{k\sqrt{T}}.
\end{aligned}
 \end{equation}
Take $m\in \left\{k,\cdots,\lfloor\frac{T}{2}\rfloor\right\}$. Summing over all $m=k,\cdots,\lfloor\frac{T}{2}\rfloor$ and rearranging, we get
\begin{equation*}
\begin{aligned}
   \mathbb{E} [V_{k}] &\leq \mathbb{E} [V_{\lfloor\frac{T}{2}\rfloor}] + \frac{1}{\sqrt{T}}\left(\frac{D^2}{2c}+cG^2\right) \sum_{m=k}^{\lfloor\frac{T}{2}\rfloor}\frac{1}{m} \\
  &\leq \mathbb{E} [V_{\lfloor\frac{T}{2}\rfloor}] + \frac{1}{\sqrt{T}}\left(\frac{D^2}{2c}+cG^2\right)\left(1+\left(\log\left(\frac{T}{2k}\right)\right)\right).   
\end{aligned}
\end{equation*}
Back to Eq.~\eqref{proof-3.1-key}, taking $w=w^{\star}$, we have
\begin{equation*}
  \mathbb{E} [V_{\lfloor\frac{T}{2}\rfloor}] \leq F(w^{\star}) + \left(\frac{D^2}{c}+2cG^2\right)\frac{1}{\sqrt{T}},
 \end{equation*}
where $V_{\lfloor\frac{T}{2}\rfloor}$ is the average value of the last $\lfloor\frac{T}{2}\rfloor$ steps, which was already analyzed in (\cite{rakhlin2011making}, Theorem 5). Then, by combining the above, we have
\begin{equation*}
\begin{aligned}
   \mathbb{E}[ F(\bar{w}_{T}^k) - F(w^{\star})] \leq & \frac{1}{k+1}\sum_{t=T-k}^{T} \mathbb{E}[ F({w}_{t}) - F(w^{\star})]+ \left(\frac{D^2}{c}+2cG^2\right)\frac{1}{\sqrt{T}}\\
  +& \frac{1}{\sqrt{T}}\left(\frac{D^2}{2c}+cG^2\right)\left(1+\left(\log\left(\frac{T}{2k}\right)\right)\right)\\
  \leq &  \frac{2+\log\left(\frac{T}{2k}\right)}{\sqrt{T}}\left(\frac{D^2}{c}+2cG^2\right).   
\end{aligned}
\end{equation*}
The first inequality follows from Jensen's inequality; we finish the proof.

\subsection{\textbf{Proof of the Theorem \ref{thm:convergence-sgd}}}\label{proof-thm-convergence-sgd} 
Let $\rho_i=1$, $\sup_{w,w^{\prime}\in {W}} \Vert w-w^{\prime} \Vert \leq D$, and pick $w=w_{T-k}$, we borrow directly from the result before Eq.~\eqref{proof-Vk} in the proof of Theorem \ref{thm:convergence-cor}. Summing over $k = \left\{1,\cdots,T-1\right\}$, we get
\begin{equation*}
\begin{aligned}
   \mathbb{E} [F_{w_T}] &=\mathbb{E}[V_0] \leq \mathbb{E} [V_{T-1}] + \frac{1}{\sqrt{T}}\left(\frac{D^2}{2c}+cG^2\right) \sum_{k=1}^{T-1}\frac{1}{k} \\
  &\leq \mathbb{E} [V_{T-1}] + \frac{1+\log{T}}{\sqrt{T}}\left(\frac{D^2}{2c}+cG^2\right),  
\end{aligned}
\end{equation*}
where $\sum_{k=1}^{T-1}\frac{1}{k} \leq 1+\log{T}$ is used in the second inequality. Back to Eq.~\eqref{proof-3.1-key}, taking $w=w^{\star}$, we have
\begin{equation*}
  \mathbb{E} [V_{T-1}] \leq F(w^{\star}) + \left(\frac{D^2}{c}+2cG^2\right)\frac{1}{\sqrt{T}}.
 \end{equation*}
then, by combining the above two inequalities, we have
\begin{equation*}
  \mathbb{E}[ F(w_{T}) - F(w^{\star})] \leq  \left(\frac{D^2}{c}+2cG^2\right)\frac{1}{\sqrt{T}}+ \frac{1+\log{T}}{\sqrt{T}}\left(\frac{D^2}{2c}+cG^2\right) 
  \leq  \frac{2+\log{T}}{\sqrt{T}}\left(\frac{D^2}{2c}+cG^2\right).
 \end{equation*}
Then, we finish the proof.

\subsection{\textbf{Proof of the Theorem \ref{thm:convergence-LAWA}}}\label{proof-thm-convergence-LAWA} 
According to the definition of LAWA in Eq.~\eqref{LAWA-rules}, we have
\begin{equation}
\begin{aligned}
    \bar{w}^{k}_{E,d} &= \frac{1}{k}\left( w_{E-k+1,d} + w_{E-k+2,d} + \cdots + w_{E,d}\right) \\
    \bar{w}^{k}_{E-1,d} &= \frac{1}{k}\left( w_{E-k,d} + w_{E-k+1,d} + \cdots + w_{E-1,d}\right).    
\end{aligned}
\end{equation}
Then, we can get recursively 
\begin{equation}
\begin{aligned}
    \bar{w}^{k}_{E,d}\! - \!\bar{w}^{k}_{E-1,d} &= \!\frac{1}{k}\left( (w_{E-k+1,d}\! -\! w_{E-k,d})\! +  (w_{E-k+2,d}\! -\! w_{E-k+1,d})\! + \! \cdots +  (w_{E,d}\! -\! w_{E-1,d})\right) \\
    &= \frac{1}{k}( \sum_{i=0}^{d-1}\alpha_{E-k+1,i} \nabla F(w_{E-k+1,i}) + \sum_{i=0}^{d-1}\alpha_{E-k+2,i} \nabla F(w_{E-k+2,i}) + \\
    &\cdots + \sum_{i=0}^{d-1}\alpha_{E,i} \nabla F(w_{E,i})), 
\end{aligned}
\end{equation}
where $\nabla F(w_{E-j,0}) = \nabla F(w_{E-j-1,d})$ and $\alpha_{E-j,0} = \alpha_{E-j-1,d}$, $j\in[0,k+1]$. Then, we have 
\begin{equation}
\begin{aligned}
    \bar{w}^{k}_{E,d} = \bar{w}^{k}_{E-1,d} - \frac{1}{k} \sum_{i=E-k+1,0}^{E,d-1}\alpha_{i} \nabla F(w_{i}).
\end{aligned}
\end{equation}
According to the definition of FWA in Eq.~\eqref{FWA-rules}, we have 
\begin{equation}
    \bar{w}_T^{k}=\frac{1}{k} \sum_{i=T-k+1}^{T} \rho_{i-(T-k)} w_{i},
\end{equation}
We slice the training steps $T$ into $E$ epochs, each containing $d$ steps, i.e., $T=Ed$. Setting $\rho_{i,d} = d$ and $\rho_{i,1} = \cdots = \rho_{i,d-1} = 0$, where $i \in [E-k+1, E]$, We can rewrite the above equality as 
\begin{equation}
    \bar{w}_{E,d}^{kd}=\frac{1}{kd}\left( d w_{E-k+1,d} + d w_{E-k+2,d} + \! \cdots + \! d w_{E,d} \right) =\frac{1}{k} \sum_{i=E-k+1,d}^{E,d}  w_{i}.
\end{equation}
Then, we have
\begin{equation}
    \bar{w}^{k}_{E,d} = \bar{w}^{k}_{E-1,d} - \frac{1}{k}\sum_{i=T-k+1,0}^{E,d-1} \alpha_i\nabla F(w_{i})= \bar{w}^{k}_{E-1,d} - \frac{1}{kd}\sum_{i=T-k+1,0}^{E,d-1} d\alpha_i\nabla F(w_{i}).
\end{equation}
The above two equality goes back to the definition of LAWA. 


Let $\sup_{w,w^{\prime}\in {W}} \Vert w-w^{\prime} \Vert \leq D$, and pick $w=w_{E-k+1,0}$, we rewrite Eq.~\eqref{proof-3.1} as
\begin{equation*}
  \mathbb{E} [\sum_{t=E-k+1,0}^{E,d-1} d(F(w_t)-F(w_{E-k+1,0})) ] 
  \leq \left(\frac{D^2}{2c}+cG^2\right)\frac{kd^2}{\sqrt{Ed}}.
 \end{equation*}
Then, let $V_{kd} = \frac{1}{kd}\sum_{t=E-k+1,0}^{E,d-1} d F({w}_{t})$, the above inequality implies that 
\begin{equation*}
  -d \mathbb{E} [F(w_{E-k+1,0})] \leq -\mathbb{E} [V_k] + \left(\frac{D^2}{2c}+cG^2\right)\frac{d}{\sqrt{Ed}}.
 \end{equation*}
According to the definition of $V_{kd}$, we get
\begin{equation}
\begin{aligned}
  \mathbb{E} [V_{kd-1}] &= \frac{1}{kd-1}(kd\mathbb{E} [V_{kd}] - d\mathbb{E} [F(w_{E-k+1,0})] )\\
  &\leq \mathbb{E} [V_{kd}] + \left(\frac{D^2}{2c}+cG^2\right)\frac{d}{(kd-1)\sqrt{Ed}}.
\end{aligned}
 \end{equation}
Take $m\in \left\{kd,\cdots,\lfloor\frac{Ed}{2}\rfloor\right\}$. Summing over all $m=kd,\cdots,\lfloor\frac{Ed}{2}\rfloor$ and rearranging, we get
\begin{equation}\label{sum-kd}
\begin{aligned}
   \mathbb{E} [V_{kd}] &\leq \mathbb{E} [V_{\lfloor\frac{Ed}{2}\rfloor}] + \frac{d}{\sqrt{Ed}}\left(\frac{D^2}{2c}+cG^2\right) \sum_{m=kd}^{\lfloor\frac{Ed}{2}\rfloor}\frac{1}{m} \\
  &\leq \mathbb{E} [V_{\lfloor\frac{Ed}{2}\rfloor}] + \frac{d}{\sqrt{Ed}}\left(\frac{D^2}{2c}+cG^2\right)\left(1+\left(\log\left(\frac{Ed}{2kd}\right)\right)\right).   
\end{aligned}
\end{equation}
Back to Eq.~\eqref{proof-3.1-key}, taking $w=w^{\star}$, we have
\begin{equation}\label{Ed/2}
  \mathbb{E} [V_{\lfloor\frac{Ed}{2}\rfloor}] \leq F(w^{\star}) + \left(\frac{D^2}{c}+2cG^2\right)\frac{d}{\sqrt{Ed}}.
 \end{equation}
then, by combining Eq.~\eqref{sum-kd} and Eq.~\eqref{Ed/2}
\begin{equation*}
\begin{aligned}
   \mathbb{E}[ F(\bar{w}_{Ed}^{k}) - F(w^{\star})] \leq &  \left(\frac{D^2}{c}+2cG^2\right)\frac{d}{\sqrt{Ed}}
  + \frac{d}{\sqrt{Ed}}\left(\frac{D^2}{2c}+cG^2\right)\left(1+\left(\log\left(\frac{Ed}{2kd}\right)\right)\right)\\
  \leq &  \frac{2d+d\log\left(\frac{E}{2k}\right)}{\sqrt{Ed}}\left(\frac{D^2}{c}+2cG^2\right).   
\end{aligned}
\end{equation*}
We finish the proof.

\vskip 0.2in
\bibliography{sample}

\end{document}